\documentclass{article}

\usepackage{amsmath,amsfonts,bm}

\DeclareMathOperator*{\argmin}{argmin} % Defines the argmin operator

\def\eqref#1{equation~\ref{#1}}
\def\1{\bm{1}}

\DeclareMathAlphabet{\mathsfit}{\encodingdefault}{\sfdefault}{m}{sl}
\SetMathAlphabet{\mathsfit}{bold}{\encodingdefault}{\sfdefault}{bx}{n}

\usepackage{microtype}
\usepackage{graphicx}
\usepackage{subcaption}
\usepackage{booktabs} % for professional tables

\usepackage{hyperref}
\usepackage{hyperref}
\usepackage{url}
\usepackage{multirow}
\usepackage{graphicx}
\usepackage{booktabs}
\usepackage{rotating}
\usepackage{makecell}
\usepackage{diagbox}
\usepackage{colortbl,xcolor} 
\usepackage{makecell} 

\newcommand{\squishlist}{
\begin{list}{{{\small{$\bullet$}}}}
{\setlength{\itemsep}{1pt}      
\setlength{\parsep}{0pt}
\setlength{\topsep}{0pt}       
\setlength{\partopsep}{0pt}
\setlength{\leftmargin}{1em} 
\setlength{\labelwidth}{1em}
\setlength{\labelsep}{0.5em} } }
\newcommand{\squishend}{  \end{list}}
\usepackage{graphicx}
\usepackage{subcaption}

\usepackage[preprint]{icml2026}

\usepackage{amsmath}
\usepackage{amssymb}
\usepackage{mathtools}
\usepackage{amsthm}

\usepackage[capitalize,noabbrev]{cleveref}

\theoremstyle{plain}

\theoremstyle{definition}

\theoremstyle{remark}

\usepackage[textsize=tiny]{todonotes}

\icmltitlerunning{UniRel: Relation-Centric Knowledge Graph Question Answering with RL-Tuned LLM Reasoning}

\begin{document}

\twocolumn[
  \icmltitle{UniRel: Relation-Centric Knowledge Graph Question Answering with RL-Tuned LLM Reasoning}



  \icmlsetsymbol{equal}{*}

  \begin{icmlauthorlist}
    \icmlauthor{Yinxu Tang}{sch}
    \icmlauthor{Chengsong Huang}{sch}
    \icmlauthor{Jiaxin Huang}{sch}
    \icmlauthor{William Yeoh}{sch}
  \end{icmlauthorlist}

  \icmlaffiliation{sch}{Washington University in St. Louis}

  \icmlcorrespondingauthor{Yinxu Tang}{t.yinxu@wustl.edu}

  \icmlkeywords{Machine Learning, ICML}

  \vskip 0.3in
]



\printAffiliationsAndNotice{}  % no special notice (required even if empty)

\begin{abstract}
 Knowledge Graph Question Answering (KGQA) has largely focused on \textit{entity-centric} queries that return a single answer entity. However, many real-world questions are inherently relational, aiming to understand how entities are associated rather than which entity satisfies a query.
In this work, we introduce \textit{relation-centric} KGQA, a complementary setting in which the answer is a subgraph that represents the semantic relations among entities. The main challenge lies in the abundance of candidate subgraphs, where trivial or overly common connections often obscure the identification of unique and informative answers. 
To tackle this, we propose \textit{UniRel}, a unified modular framework that combines a subgraph retriever with an LLM fine-tuned using reinforcement learning. The framework uses a reward function to prefer compact and specific subgraphs with informative relations and low-degree intermediate entities.
Experiments show that UniRel improves connectivity and reward over Prompting baselines and generalizes well to unseen entities and relations. Moreover, UniRel can be applied to conventional entity-centric KGQA, achieving competitive or improved performance in several settings.

% Knowledge Graph Question Answering (KGQA) has traditionally focused on \textit{entity-centric} queries that return a single answer entity. 
% However, real-world queries are often relational, seeking to understand how entities are associated.
% In this work, we introduce \textit{relation-centric} KGQA, a complementary setting where the answer is a subgraph capturing the semantic connections among entities rather than an individual entity. 
% The main challenge lies in the abundance of candidate subgraphs, where trivial or overly common connections often obscure the identification of unique and informative answers. 
% To tackle this, we propose \textit{UniRel-R1}, a unified framework that integrates subgraph selection, multi-stage graph pruning, and an LLM fine-tuned with reinforcement learning. The reward function is designed to encourage compact and specific subgraphs with more informative relations and lower-degree intermediate entities.
% Extensive experiments show that UniRel-R1 achieves significant gains in connectivity and reward over Prompting baselines and generalizes effectively to unseen entities and relations.

% We propose a new framework XXX, which includes three modules: subgraph selection, graph pruning and a tuned LLM. We use reinforcement learning to finetune the LLM to better understand how informative a path is. Empircal results show that xxx can achieve xxxx. Further analysis prove that our method can generalize to unseen tasks.
\end{abstract}
\section{Introduction}
\begin{figure*}[t]
    \centering
    \includegraphics[width=0.92\textwidth]{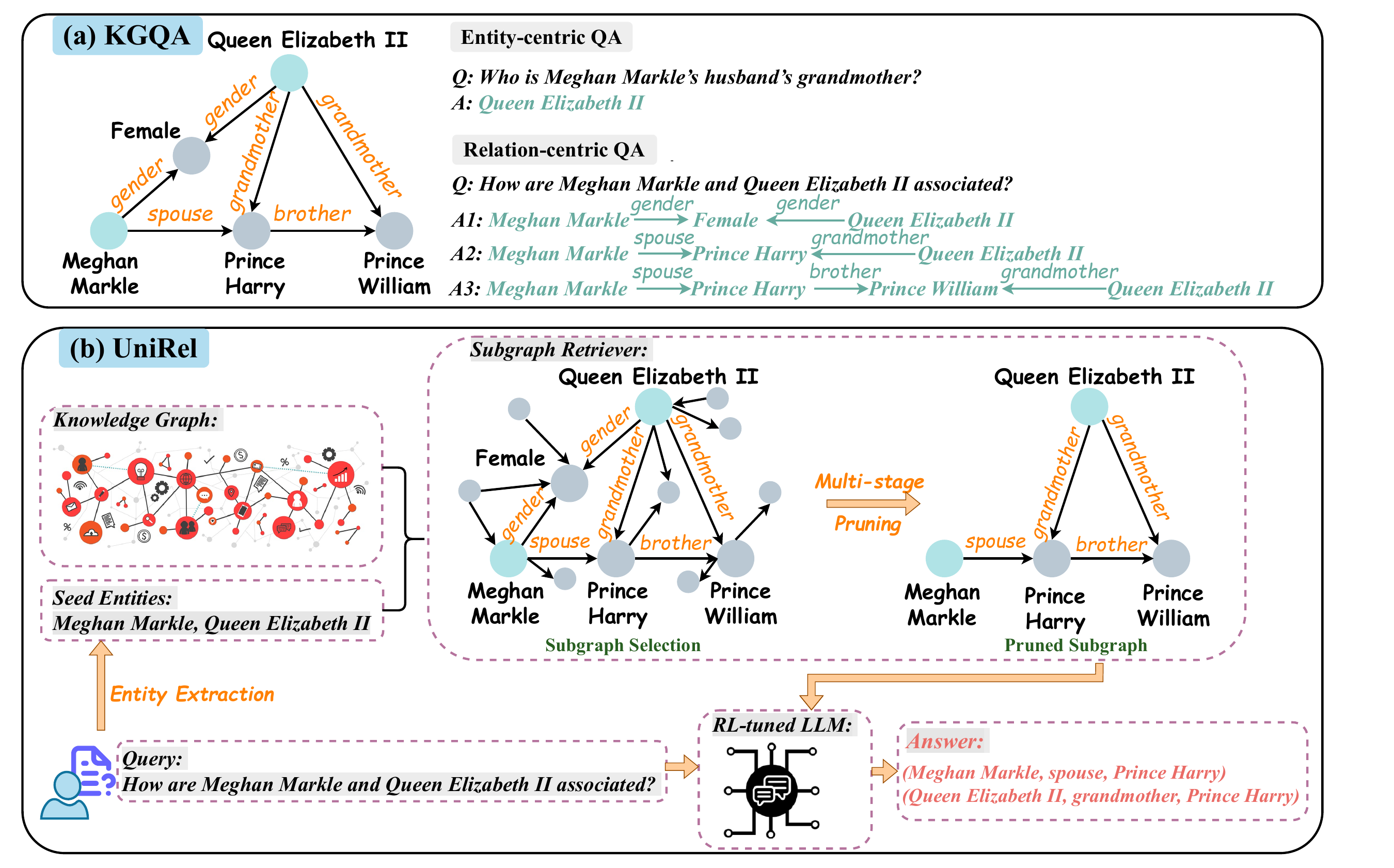}
    \vspace{-0.7em}
    \caption{An overview of KGQA and our proposed UniRel framework. (a) An illustrative example contrasting entity-centric KGQA with relation-centric KGQA. (b) The UniRel pipeline, integrating subgraph retriever and RL-tuned LLM reasoning.}
    \vspace{-1.0em}
    \label{fig:overview}
\end{figure*}

Traditional knowledge graph question answering (KGQA) is largely \textit{entity-centric}, aiming to return a single target entity.
For example, a query such as ``Who is Meghan Markle’s husband’s grandmother?'' yields ``Queen Elizabeth II'' as the answer (see Figure~\ref{fig:overview}(a)).
Alternatively, users also often pose \textit{relation-centric} queries that seek to uncover the relationships between entities rather than retrieve an individual fact. For example, ``How are Meghan Markle and Queen Elizabeth II associated?'' requires constructing a subgraph that captures their semantic connections. Such queries are common in real-world settings but remain beyond the scope of traditional KGQA systems.

Existing work in KGQA has primarily adhered to the entity-centric paradigm, often augmenting LLMs with structured knowledge to improve accuracy.
Retrieval-augmented generation (RAG) incorporates KG facts as contextual input~\citep{linders2025knowledge}, while other approaches combine LLMs with graph neural networks (GNNs) for joint reasoning over text and structure~\citep{xu2025harnessing,yasunaga2021qa,he2024g}.
To enhance interpretability, some methods generate explicit reasoning paths for multi-hop queries~\citep{zhou2018interpretable,chakraborty2024multi,zhang2018variational}, and others employ reinforcement learning (RL) or search strategies such as Monte Carlo Tree Search (MCTS) to explore candidate paths~\citep{shen2025reasoning}. Despite these advances, current systems ultimately return a single entity, leaving relation-centric queries unexplored.

In this work, we introduce \textit{relation-centric} KGQA, a complementary setting to the standard \textit{entity-centric} KGQA. Instead of producing an entity, the answer is a subgraph that captures the underlying relational structure among the seed entities. 
As demonstrated in Figure~\ref{fig:overview}(a), multiple subgraphs may serve as answers. While Answer~A1 is the shortest, it reflects an overly common relation and conveys little information, whereas Answers~A2 and~A3 provide more unique and informative connections.
This illustrates the central challenge of relation-centric KGQA: Among many candidate subgraphs, trivial or generic ones often obscure unique and informative answers.

To address this challenge, we propose \textit{UniRel}, a unified modular framework for KGQA, with a particular focus on relation-centric queries.
UniRel integrates \textit{subgraph retriever} and \textit{RL-tuned LLM} reasoning into a coherent pipeline.
Figure~\ref{fig:overview}(b) illustrates the overall workflow.
Starting from a natural language query, UniRel first identifies seed entities and then applies a subgraph retriever module that combines candidate selection with multi-stage pruning, producing a compact and task-relevant subgraph.
The retrieved subgraph is textualized and provided alongside the query to an RL-tuned LLM, where the reward design encourages concise and informative outputs by favoring smaller structures with rarer relations and lower-degree intermediates.
Overall, UniRel produces answers that are not only valid but also capture distinctive and meaningful relations.

% To address this challenge, we propose \textit{UniRel}, a unified modular framework for relation-centric KGQA.
% UniRel-R1 integrates \textit{subgraph selection}, \textit{multi-stage graph pruning}, and \textit{RL-tuned LLM}. 
% Figure~\ref{fig:overview}(b) illustrates the pipeline, which begins by extracting seed entities from a natural language query.
% It then performs subgraph selection to identify candidate relational structures.
% A pruning stage follows to remove trivial or overly generic entities and relations, yielding a compact subgraph.
% The subgraph is textualized and provided alongside the query to an RL-tuned LLM, where the reward design encourages concise and informative outputs by favoring smaller structures with rarer relations and lower-degree intermediates.
% In this way, UniRel ensures that the final answers are not only valid but also capture distinctive and meaningful relations.

We conduct extensive experiments across seven benchmark knowledge graphs and a diverse set of LLMs from the Qwen and Llama families.
To this end, we build relation-centric query sets covering both two-entity and multi-entity settings, evaluated under in-domain and cross-domain scenarios.
Results show that \textit{UniRel} consistently outperforms Prompting baselines, yielding at least a \textbf{35\%} improvement in connectivity and over a \textbf{245\%} gain in reward, while generalizing to unseen entities and relations.
Interestingly, Qwen models are substantially more sensitive to the removal of semantic information than Llama models, highlighting a fundamental difference in how these families leverage semantic cues. 
Moreover, UniRel can also generalize to conventional entity-centric KGQA by replacing the subgraph retriever module with an entity-centric variant, achieving \textit{competitive} or \textit{improved} performance compared to the existing baselines.
% The contributions of this paper are outlined as follows:
% \begin{itemize}
%     \item \textbf{Introducing Relation-Centric QA.} We formalize a complementary QA setting where answers are subgraphs capturing semantic associations among entities, beyond the traditional entity-centric paradigm.  
%     \item \textbf{Developing the UniRel-R1 Framework.}  We propose a unified pipeline that integrates subgraph selection, graph pruning, and reinforcement learning–based LLM fine-tuning. Our principled reward design explicitly encourages valid, unique, and informative answers.  
%     \item \textbf{Constructing Benchmarks for Evaluation.} We build relation-centric query sets across seven KG datasets, including extensions to multi-entity queries, providing a systematic testbed for this new task.  
%     \item \textbf{Empirical Evaluation and Findings.} Through extensive experiments on Qwen and Llama models, we show that UniRel-R1 achieves significant gains in connectivity and informativeness over zero-shot baselines, while generalizing to unseen entities and relations. Moreover, our analysis reveals that Qwen models are more sensitive to the removal of semantic information compared with Llama models.  
% \end{itemize}

\section{Problem Statement}
In relation-centric KGQA, a knowledge graph is formally defined as $\mathcal{G} = (\mathcal{E}, \mathcal{R}, \mathcal{T})$, where $\mathcal{E}$ is the set of entities, $\mathcal{R}$ the set of relation types, and
$\mathcal{T} \subseteq \mathcal{E} \times \mathcal{R} \times \mathcal{E}$ the set of triples. Given a query $q$, let $\mathcal{E}_q \subseteq \mathcal{E}$ denote the set of seed entities extracted from $q$.

The task is to construct a subgraph: 
\small
$$\mathcal{G}^* = (\mathcal{E}^*, \mathcal{R}^*, \mathcal{T}^*) \subseteq \mathcal{G}.$$
\normalsize
A subgraph $\mathcal{G}^*$ is considered \textit{valid} if it connects all entities in $\mathcal{E}_q$, with $\mathcal{E}_q \subseteq \mathcal{E}^*$, thereby ensuring that the answer is grounded in the semantics of $\mathcal{G}$. 
When multiple valid subgraphs exist, user preferences may vary depending on their background knowledge or task requirements.
In this work, we prioritize subgraphs that are more \textit{informative}, rather than those dominated by trivial or overly common structures.
\textit{Accordingly, our optimization objective balances compactness, relation rarity, and lower-degree intermediate entities}, ensuring that the resulting subgraphs emphasize non-trivial and distinctive semantic relations.
Alternative optimization objectives can also be defined, depending on the specific application scenario.

\section{Framework: UniRel}

This section outlines the two-stage \textit{UniRel} pipeline.
Given a query, UniRel first performs \textit{subgraph retriever} to construct and refine a task-specific subgraph via multi-stage pruning (Secs.~\ref{sec:subgraph} and~\ref{sec:prune}).
In the second stage, an RL-tuned LLM reasons over the retrieved subgraph and the query to generate an answer emphasizing the most informative relations (Sec.~\ref{sec:llm_training}). Note that the framework can be extended to alternative optimization objectives by modifying the pruning criteria and adjusting the reward function.

\begin{figure*}[t]
    \centering
    \includegraphics[width=0.86\textwidth]{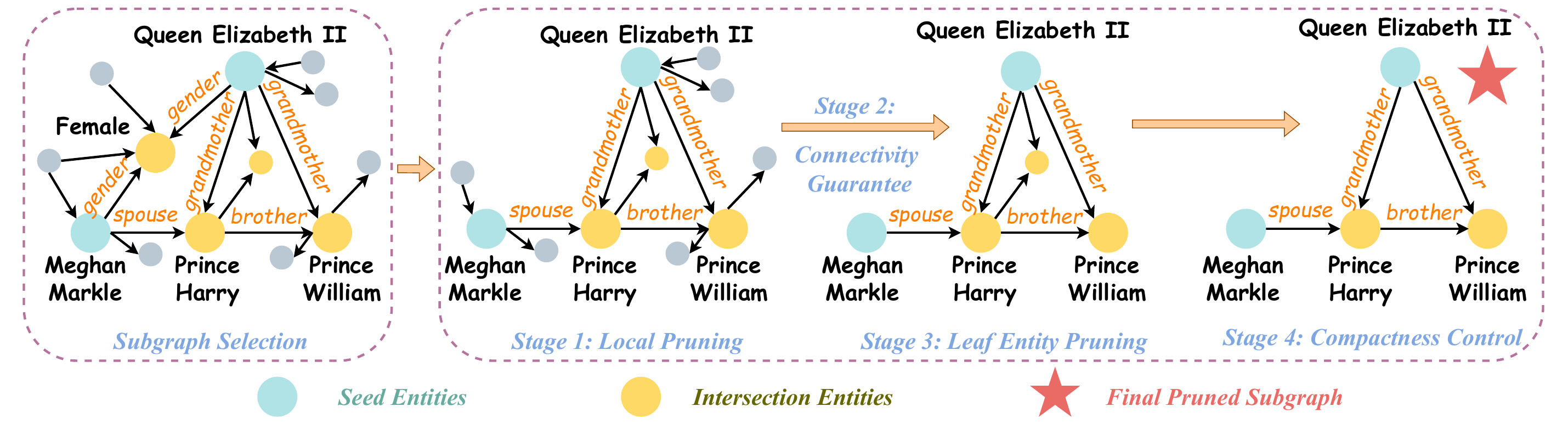}
    \vspace{-0.5em}
    \caption{An example of multi-step graph pruning. From the initial subgraph (left), Stage~1 removes overly common entities (e.g., \textit{Female}), Stage~2 enforces connectivity through intersections, Stage~3 prunes leaf entities, and Stage~4 enhances compactness by removing high-penalty intersection nodes (yellow). The final subgraph (right) retains the most informative relations among the seed entities.}
    \vspace{-1em}
    \label{fig:graph_pruning}
\end{figure*}

% In this section, we outline the overall pipeline of \textit{UniRel-R1}, which consists of three stages.
% A task-specific subgraph is first constructed from the knowledge graph based on entities mentioned in the query (Sec.~\ref{sec:subgraph}).
% The subgraph is then refined through multi-stage pruning to remove redundant or trivial entities and relations (Sec.~\ref{sec:prune}).
% Finally, an RL-tuned LLM processes the textualized subgraph with the query to generate an answer that highlights the most informative relations(Sec.~\ref{sec:llm_training}).\footnote{Note that the framework can be extended to alternative optimization objectives by modifying the pruning criteria and adjusting the reward function accordingly.}
\subsection{Subgraph Retriever}
Subgraph retriever aims to construct a compact, task-relevant subgraph for downstream reasoning and consists of two components: subgraph selection and multi-step pruning. More details are shown in Appendix~\ref{sec:complexity}.
\subsubsection{Subgraph Selection}
\label{sec:subgraph}

Given a query $q$, the seed entity set is extracted from $q$ as $\mathcal{E}_q = \{e_1, e_2, \dots, e_m\}$.
For each seed entity $e_i \in \mathcal{E}_q$, we define its $k$-hop expansion as
{\small
$
\mathcal{V}^k(e_i) = \{ v \in \mathcal{E} \mid \mathrm{dist}(e_i,v) \leq k \},
\mathcal{T}^k(e_i) = \{ (u,r,v) \in \mathcal{T} \mid u,v \in \mathcal{V}^k(e_i) \},
$}
where $\mathrm{dist}(\cdot,\cdot)$ denotes the distance in the undirected projection of $\mathcal{G}$.

The candidate subgraph $\mathcal{G}' = (\mathcal{V}', \mathcal{R}', \mathcal{T}')$ is then obtained by aggregating the $k$-hop neighborhoods of all seed entities, where 
\small
$
\mathcal{V}' = \bigcup\nolimits_{e_i \in \mathcal{E}_q} \mathcal{V}^k(e_i), 
\mathcal{T}' = \bigcup\nolimits_{e_i \in \mathcal{E}_q} \mathcal{T}^k(e_i),
$
\normalsize
and
\small
$
\mathcal{R}' = \{ r \in \mathcal{R} \mid (u,r,v) \in \mathcal{T}' \}.
$
\normalsize
The subgraph $\mathcal{G}'$ serves as the initial candidate, refined through pruning in the next stage.

\subsubsection{Multi-step Graph Pruning}
\label{sec:prune}
The candidate subgraph $\mathcal{G}' = (\mathcal{V}', \mathcal{R}', \mathcal{T}')$  
often contains redundant structures and overly generic entities that obscure meaningful relational patterns.  
To mitigate this issue, we refine $\mathcal{G}'$ through a principled multi-stage pruning process.  

As a measure of entity generality, we introduce the \textit{hub penalty}, defined for each $e \in \mathcal{E}$ as:
{\small
\begin{equation}\label{eq:hub_penalty}
    \mathrm{HubPenalty}(e) = \log \big( 1 + \deg(e) \big),
\end{equation}
}
where $\deg(e)$ denotes the degree of $e$ in the undirected projection of $\mathcal{G}$.  
The penalty increases with entity degree, such that highly connected entities (e.g., \textit{male}) 
receive larger values, while rarer and more specific entities (e.g., \textit{president}) are penalized less.
This formulation provides a principled criterion for filtering out uninformative entities and guiding the pruning of $\mathcal{G}'$.

Figure~\ref{fig:graph_pruning} shows the multi-stage pruning on the example in Figure~\ref{fig:overview}, with each stage explained below.
% \paragraph{Stage 1: Local Pruning.}  

\textbf{Stage 1: Local Pruning.}
For each expansion set $\mathcal{V}^k(e_i)$, we impose a threshold $\rho$ on the hub penalty to eliminate overly common entities.    
Each \textbf{non-seed}  $v$ with $\mathrm{HubPenalty}(v) \geq \rho$ is removed along with its incident relations.  
Subsequently, entities that become isolated (i.e., with no incident relations) are discarded.  
The resulting pruned neighborhoods are denoted by $\mathcal{V}_\rho^k(e_i)$ and $\mathcal{T}_\rho^k(e_i)$.
% \paragraph{Stage 2: Connectivity Guarantee.} 

\textbf{Stage 2: Connectivity Guarantee.}
Following local pruning, the filtered neighborhoods $\{\mathcal{V}_\rho^k(e_i)\}$ may fail to ensure connectivity among all seed entities.  
To evaluate connectivity, we compute pairwise intersections 
\small
$\mathcal{I}_{ij} = \mathcal{V}_\rho^k(e_i) \cap \mathcal{V}_\rho^k(e_j)$ 
\normalsize
and construct an auxiliary graph
\small
$
\mathcal{H}_\rho = ( \{ \mathcal{V}_\rho^k(e_i) \}, \; \{ (i,j) \mid \mathcal{I}_{ij}\neq \emptyset \} ),
$
\normalsize
where each node represents a pruned neighborhood $\mathcal{V}_\rho^k(e_i)$,  
and an edge $(i,j)$ is introduced whenever their intersection is non-empty.  
If $\mathcal{H}_\rho$ is connected, then all seed entities are jointly connected in the candidate subgraph.  
Otherwise, $\rho$ is incrementally relaxed and Stages~1--2 are repeated until graph $\mathcal G_\rho = (\mathcal{V}_{\rho}, \mathcal{R}_{\rho}, \mathcal{T}_{\rho})$ becomes connected.

% Otherwise, the threshold $\rho$ is incrementally relaxed and Stage~1 is reapplied until connectivity is restored.
% Otherwise, the threshold $\rho$ is incrementally relaxed, and Stages~1--2 are reapplied to obtain an updated subgraph $\mathcal G'_\rho$ until connectivity is restored.

\textbf{Stage 3: Leaf Entity Pruning.}
% \paragraph{Stage 3: Leaf Entity Pruning.}  
Once connectivity is ensured, we iteratively eliminate non-seed entities in  $\mathcal G_\rho$ whose neighbor set \small $\mathcal{N}_{\rho}(v) = \{ u \in \mathcal{V}_{\rho} \mid \exists r \in \mathcal{R}_{\rho}: (v,r,u) \in \mathcal{T}_{\rho}$ or $(u,r,v) \in \mathcal{T}_{\rho} \}$ \normalsize contains only a single distinct element (\small $|\mathcal{N}_{\rho}(v)| = 1$\normalsize). Such leaf-like entities and their incident relations are iteratively pruned until no removal is possible. 

% Once connectivity is ensured, we iteratively remove non-seed entities in $\mathcal G_\rho$ whose neighbor set
% \[
% \mathcal N(v) = \{ u \in \mathcal V_\rho \mid \exists r \in \mathcal R_\rho : (v,r,u) \in \mathcal T_\rho \ \text{or}\ (u,r,v) \in \mathcal T_\rho \}
% \]
% contains only a single distinct element, i.e., $|\mathcal N(v)| = 1$.

\textbf{Stage 4: Compactness Control.}
% \paragraph{Stage 4: Compactness Control.} 
If the refined subgraph remains large $\mathcal G'_\rho$, we further enforce compactness by leveraging the intersections \small $\{\mathcal{I}_{ij}\}$ \normalsize defined in Stage~2.  
Let \small $\mathcal{I}^+ = \{\mathcal{I}_{ij} \mid \mathcal{I}_{ij} \neq \emptyset\}$ \normalsize denote the collection of all non-empty intersections.  
We then select a subset \small $\mathcal{I}^c \subseteq \mathcal{I}^+$ \normalsize whose index pairs jointly cover all seed entities, i.e.,
\small
$
\bigcup\nolimits_{\mathcal{I}_{ij} \in \mathcal{I}^c} \{i,j\} = \{0,1,\dots,m-1\}.
$
\normalsize
For each valid subset $\mathcal{I}^c$, a candidate subgraph is constructed in two steps.  
First, within each intersection $\mathcal{I}_{ij}$, we retain $s$ entities with lowest hub penalties:
$$
\widehat{\mathcal{I}}_{ij} 
= \argmin\nolimits_{\substack{\mathcal{U} \subseteq \mathcal{I}_{ij}, |\mathcal{U}| = s}} 
\sum\nolimits_{v \in \mathcal{U}} \mathrm{HubPenalty}(v).
$$
Second, each selected set $\widehat{\mathcal{I}}_{ij}$ is expanded via a $k$-hop search restricted to the corresponding pruned neighborhoods:
\small
$
\mathcal{N}^k(\widehat{\mathcal{I}}_{ij}) 
= \widehat{\mathcal{I}}_{ij} \cup \{ u \in \mathcal{V}_\rho^k(e_i) \cup \mathcal{V}_\rho^k(e_j) \mid \mathrm{dist}(u,v) \leq k ,\ v \in \widehat{\mathcal{I}}_{ij} \}.
$
\normalsize
The candidate node set associated with $\mathcal{I}^c$ is then
\small $\mathcal{V}(\mathcal{I}^c) = \bigcup_{\mathcal{I}_{ij} \in \mathcal{I}^c} \mathcal{N}^k(\widehat{\mathcal{I}}_{ij})$\normalsize,
and its induced subgraph is subsequently simplified using Stage~3.  

Finally, the most compact subgraph under parameters $(s,\rho)$ is obtained as:
\small
\[
\mathcal{G}^{s}_{\rho} = (\mathcal{V}^{s}_{\rho}, \mathcal{R}^{s}_{\rho}, \mathcal{T}^{s}_{\rho}),
\]
\normalsize
where \small $\mathcal{V}^{s}_{\rho} = \argmin_{\mathcal{S} \in \mathcal{I}^+} \; |\mathcal{V}(\mathcal{S})|$, $\mathcal{T}^{s}_{\rho} \!= \!\{ (u,r,v) \in \mathcal{T}'_{\rho} \mid u,v \in \mathcal{V}^{m}_{\rho} \}$, \small $\mathcal{R}^{s}_{\rho} = \{ r \in \mathcal{R}'_{\rho} \mid (u,r,v) \in \mathcal{T}^{s}_{\rho} \}$\normalsize.

\textbf{Stage 5: Iterative Reduction.}
% \paragraph{Stage 5: Iterative Reduction.}   
If the subgraph still exceeds the target size, we decrease $s$ and reapply Stage~4
until a sufficiently compact subgraph is obtained.

\subsection{Relational Answer Generation via LLM}
\label{sec:llm_training}
Given the refined subgraph $\mathcal{G}^{m}_{\rho}$, the final step is to generate an answer that highlights the most informative relational structure.  
We employ an RL-tuned LLM to produce relational explanations conditioned on the query and textualized subgraph,
optimized using task-specific rewards detailed in the following sections.

\subsubsection{Reinforcement Learning Algorithms}
\textbf{RL with Verifiable Reward (RLVR).}
Our approach builds on the RLVR paradigm~\citep{Lambert2024TLU3P}, which applies to settings where response quality can be deterministically verified.
In its standard form, RLVR uses a rule-based verifier $v : \mathcal{X} \to \{0,1\}$ to assign binary rewards: $r_i = v(x_i)$, where $v(x_i)=1$ if $x_i$ passes a task-specific correctness check, and $0$ otherwise.
This binary scheme is effective for tasks with unambiguous success criteria (e.g., mathematical problem-solving and code generation).  
In our setting, we design a corresponding rule-based reward to verify model generated relations, described in Sec.~\ref{sec:reward}.

\textbf{Group Relative Policy Optimization (GRPO).}
For policy optimization, we adopt GRPO~\citep{shao2024deepseekmath},  
which evaluates responses based on their relative performance within a sampled group from the policy.  

Given a query $q$, we sample from the previous policy $\pi_{\theta_{\mathrm{old}}}(\cdot\!\mid q)$ to get $G$ responses $\{o_i\}_{i=1}^G$ with rewards $\{r_i\}$.
The advantage $A_i$ is the group-normalized reward, defined as:
\small
$$
A_i = \frac{r_i - \overline{r}}
     {\sqrt{\tfrac{1}{G}\sum_{j=1}^G (r_j - \overline{r})^2} + \varepsilon_{\mathrm{stab}}}, \quad \overline{r} = \frac{1}{G}\sum_{j=1}^G r_j,
$$
\normalsize
where $\varepsilon_{\mathrm{stab}}$ ensures numerical stability. 
With $\rho_i=\frac{\pi_{\theta}(o_i\!\mid q)}{\pi_{\theta_{\mathrm{old}}}(o_i\!\mid q)}$,
% the GRPO objective maximizes a clipped surrogate similar to PPO:
GRPO loss is defined as a clipped surrogate similar to PPO:
% ~\jh{I feel that you can use J as the symbol for training objective (as the one used in GRPO paper) instead of L, which means loss function (which needs a negative sign here).}
\small \begin{equation*}
\begin{aligned}
    \mathcal{L}_{\mathrm{GRPO}}(\theta)
 =&
-
[
\frac{1}{G}\sum_{i=1}^G
\min\!\bigl(\rho_i A_i,\;\!\!\mathrm{clip}(\rho_i,1-\alpha,1+\alpha)\,A_i\bigr) \\
& -\;\!\!\beta\,D_{\mathrm{KL}}\bigl(\pi_{\theta}(\cdot|q)\,\Vert\,\pi_{\mathrm{ref}}(\cdot|q)\bigr)
].
\end{aligned}    
\end{equation*}
\normalsize
This objective encourages policy $\pi_{\theta}$ to favor above-average responses.
A $\beta$-scaled KL term regularizes updates to limit deviations from the reference policy $\pi_{\mathrm{ref}}$.

\normalsize
\subsubsection{Reward design}
\label{sec:reward}

In relation-centric KGQA, a valid answer must follow the format, ensure connectivity, and emphasize informative entities and relations.  
To capture these aspects, we design a composite reward with four components: \emph{Format}, \emph{Connectivity}, \emph{Entity Informativeness}, and \emph{Relation Informativeness}.

Formally, for each answer $a$, we define:

\textbf{Format Reward.} The format reward is defined as $R_{\mathrm{fmt}}(a) \in \{-1,1\}$, where $R_{\mathrm{fmt}}(a) = 1$ when $a$ conforms to the required format and $R_{\mathrm{fmt}}(a) = -1$ otherwise.

\textbf{Connectivity Reward.}
% \paragraph{Connectivity Reward.} 
The connectivity reward evaluates whether the output subgraph connects the extracted entities $\mathcal{E}_q$ and is defined as $R_{\mathrm{con}}(a) \in \{-\lfloor |\mathcal{E}_q|/2 \rfloor, \dots, \lceil |\mathcal{E}_q|/2 \rceil - 1\}$, where the minimum indicates full disconnection, intermediate values reflect partial connectivity (e.g., $-\lfloor |\mathcal{E}_q|/2 \rfloor + l$ for $(l+1)$-connected entities), and the maximum denotes full connectivity.

% The connectivity reward evaluates whether the output subgraph connects the extracted entities $\mathcal{E}_q$, with the following cases:  
% \squishlist
%  \item $|\mathcal{E}_q| = 2$: $R_{\mathrm{rch}}(a) \in \{-1,0\}$, taking value $0$ if the two entities are connected and $-1$ otherwise. 
%   \item$|\mathcal{E}_q| > 2$: $R_{\mathrm{rch}}(a) \in \{-\lceil |\mathcal{E}_q|/2 \rceil + 1, \dots, \lfloor |\mathcal{E}_q|/2 \rfloor\}$, where the minimum indicates full disconnection, intermediate values reflect partial connectivity (e.g., $-\lceil |\mathcal{E}_q|/2 \rceil + k$ for $k$-connected entities), and the maximum denotes full connectivity.
% \squishend

\textbf{Entity Informativeness Reward.} 
This component favors informative over generic entities.  
For a generated answer $a$ with entity set $\mathcal{E}(a)$,  
the reward is the sum of normalized hub penalties:
\vspace{-1em}
\small
\begin{equation}\label{eq:entity_informativenss}
R_{\mathrm{ent}}(a) = \!\!\!\!\sum_{e \in \mathcal{E}(a)} \!\!\!R_{\mathrm{ent}}(e), R_{\mathrm{ent}}(e) \!= \!\!- \frac{\mathrm{HubPenalty}(e)}{\max_{v \in \mathcal{E}} \mathrm{HubPenalty}(v)},
\end{equation}
\normalsize
where $\mathrm{HubPenalty}(e)$ is defined in Equation~\ref{eq:hub_penalty}.

\textbf{Relation Informativeness Reward.} This component favors infrequent relations, which are more informative.
The informativeness of a relation $r$ is measured by its inverse document frequency (IDF)~\citep{robertson2004understanding}:
\small
$
\mathrm{IDF}(r) = \log \!\left(\frac{|\mathcal{T}|}{|\{(u,r,v)\in \mathcal{T}\}|}\right),
$
\normalsize
where $|\mathcal{T}|$ is the number of triples in $\mathcal{G}$ and the denominator is the frequency of $r$.  
Frequent relations (e.g., \textit{gender}) receive lower IDF scores, while rarer ones (e.g., \textit{invention}) score higher. 

Given an answer $a$ with relation set $\mathcal{R}(a)$, the reward is the sum of normalized IDF scores:
\small
\begin{equation}\label{eq:relation_informativenss}
\vspace{-1em}
R_{\mathrm{rel}}(a) = \sum_{r \in \mathcal{R}(a)} R_{\mathrm{rel}}(r),
R_{\mathrm{rel}}(r) = \frac{\mathrm{IDF}(r)}{\max\nolimits_{s \in \mathcal{R}} \mathrm{IDF}(s)} - 1.
\end{equation}
\normalsize

\textbf{Overall Reward.}  
The final reward for an answer $a$ aggregates all four components:
\small
\begin{equation}\label{eq:reward}
    R(a) = R_{\mathrm{fmt}}(a) 
+ R_{\mathrm{con}}(a) 
+ \tfrac{1}{2}\!\left(\tfrac{R_{\mathrm{ent}}(a)}{x} 
+ \tfrac{R_{\mathrm{rel}}(a)}{y}\right),
\end{equation}
\normalsize
where $x$ and $y$ are normalization constants ensuring  
$\tfrac{R_{\mathrm{ent}}(a)}{x}, \tfrac{R_{\mathrm{rel}}(a)}{y} \in [-1,0]$,  
with values determined by the maximum permitted subgraph size.  

By construction, the reward is bounded as a function of the number of seed entities $|\mathcal{E}_q|$: $R(a) \in [-\lfloor |\mathcal{E}_q|/2 \rfloor- 2, \lceil |\mathcal{E}_q|/2 \rceil)$.
If $R_{\mathrm{fmt}}(a)=-1$, then $R(a)=-\lfloor |\mathcal{E}_q|/2 \rfloor- 2$.  
If $R_{\mathrm{con}}(a)=-\lfloor |\mathcal{E}_q|/2 \rfloor$,  
then $R(a)=-\lfloor |\mathcal{E}_q|/2 \rfloor$.

% \small
% \[
% R(a) \in 
% \begin{cases}
% [-3,\,1), & \text{if } |\mathcal{E}_q| = 2, \\[6pt]
% \bigl[-\lceil |\mathcal{E}_q|/2 \rceil - 1,\;\lfloor |\mathcal{E}_q|/2 \rfloor + 1\bigr), & \text{if } |\mathcal{E}_q| > 2.
% \end{cases}
% \]
% \normalsize
% Let $R_{\min}(|\mathcal{E}_q|)$ denote the lower bound.  
% If $R_{\mathrm{fmt}}(a)=-1$, then $R(a)=R_{\min}(|\mathcal{E}_q|)$.  
% If $R_{\mathrm{con}}(a)=-\lceil |\mathcal{E}_q|/2 \rceil+1$,  
% then $R(a)=R_{\min}(|\mathcal{E}_q|)+2$.

\section{Experiment}

In this section, we evaluate the effectiveness of \textit{UniRel} on relation-centric KGQA across multiple datasets and LLMs.
We first describe the datasets and models, then present the main results, followed by analyses of generalization and scalability under multi-entity queries.
Finally, we apply the \textit{UniRel} pipeline to conventional entity-centric KGQA to assess its transferability.

\subsection{Experiment Setup}

\textbf{Datasets.} We evaluate on seven benchmark KG datasets to ensure diversity in scale, domain, and relational complexity:
Freebase13~\citep{Socher2013ReasoningWN}, FB15k-237~\citep{toutanova2015observed}, MetaQA~\citep{zhang2018variational},
DBpedia50/500~\citep{Auer07DBpedia}, YAGO3-10~\citep{Mahdisoltani2015YAGO3AK}, and UMLS~\citep{bodenreider2004umls}.
These datasets span encyclopedic, biomedical, and commonsense knowledge, providing a testbed for relation-centric KGQA.
Dataset statistics are reported in Appendix~\ref{sec:implement_details}.

\textbf{Query Construction.} Existing KGQA benchmarks focus on \textit{entity-centric} queries that return a single entity.  
To adapt them to the relation-centric setting, we constructed 2,500 queries per dataset, with 2,000 for training and 500 for testing.  
For each dataset, we generated two-entity queries with $\mathrm{dist}(e_i, e_j) \leq 4$ in the knowledge graph.  
To evaluate the scalability of \textit{UniRel-R1}, we further created multi-entity queries (involving three and four entities) on DBpedia50. 
Full construction details are provided in Appendix~\ref{sec:implement_details}.

\textbf{Models.}
To assess generalizability of our method, we evaluated \textit{UniRel} on LLMs from the Qwen~\citep{yang2025qwen3} and Llama~\citep{grattafiori2024llama} families, covering diverse architectures and scales.  
Specifically, we used Qwen-2.5-3B/7B/14B-Instruct, Llama-3.2-3B-Instruct, and Llama-3.1-8B-Instruct, spanning 3B–14B parameters. 
%These results demonstrate the broad applicability and robustness of our approach across different model sizes and architectures.

\textbf{Prompts.} 
We designed structured prompts that encode the textualized knowledge graph as node and edge tables and incorporate the query, requiring the model to return the corresponding subgraph in a standardized triple format.
This ensures parsable outputs that are directly comparable across models.
Prompt templates are provided in Appendix~\ref{sec:implement_details}.

\textbf{Baselines.} We compare UniRel with a \textit{Prompting} baseline that shares the same overall pipeline, differing only in the final generation stage.
Specifically, Prompting uses an LLM without RL fine-tuning to generate relational explanations.

\textbf{Evaluation Metrics.} 
We evaluate performance using three primary metrics: \emph{Connectivity Ratio} ($C$), \emph{Average Reward} ($\bar{R}$), and \emph{Subgraph F1}.
The \emph{Connectivity Ratio} measures whether the generated subgraph successfully connects all seed entities, while the \emph{Average Reward} reflects the overall subgraph quality as defined in Equation~\ref{eq:reward}.
\emph{Subgraph F1} is a structure-level analogue of evidence-level F1 used in explainable multi-hop QA, measuring how well the predicted subgraph overlaps with a reference reasoning subgraph in terms of both coverage and compactness. Its definition and
additional implementation details are in Appendix~\ref{sec:implement_details}.

\subsection{Main Results} 

\textbf{Parameter Choice.} 
For the normalization terms $x$ and $y$ in Equation~\ref{eq:reward}, we calibrate their values using the Qwen-2.5-3B-Instruct model on the largest dataset, DBpedia500.
Importantly, the choice of $x$ and $y$ primarily depends on the query construction protocol rather than dataset-specific statistics.
Since query construction is consistent across all datasets, we set $x=7$ and $y=6$ and apply these values uniformly across all datasets and models to ensure consistency and comparability.
Further details of this parameter tuning are provided in Appendix~\ref{sec:main_results_details}.
\begin{table*}[t]
\centering
\caption{Performance comparison between Prompting and UniRel, reported as (connectivity ratio, average reward, subgraph F1).} 
\resizebox{\textwidth}{!}{%
\begin{tabular}{llccccccc}
\toprule
\multicolumn{2}{c}{\diagbox{\textbf{Models}}{\textbf{Datasets}}} & \textbf{Freebase13} & \textbf{FB15k-237} & \textbf{MetaQA} & \textbf{DBpedia50} & \textbf{DBpedia500} & \textbf{YAGO3-10} & \textbf{UMLS} \\
\midrule
\multirow{2}{*}{\textbf{Qwen3B}}          
& Prompting    & (41.2\%, -0.53, 29.3) & (0.6\%, -1.41, 0.5)  & (0.4\%, -1.30, 5.0)  & (11.2\%, -1.10, 15.5) & (0.6\%, -1.43, 3.5)  & (1.0\%, -1.4, 4.1)   & (23.0\%, -0.73, 11.7) \\
& UniRel  & (86.0\%, 0.64, 58.7)  & (62.0\%, -0.02, 2.7) & (72.8\%, 0.07, 25.0)  & (91.0\%, 0.55, 39.2)  & (64.8\%, -0.06, 14.3) & (67.4\%, -0.03, 15.8) & (69.0\%, 0.08, 17.1)  \\ 
\multirow{2}{*}{\textbf{Qwen7B}}          
& Prompting    & (57.8\%, 0.04, 42.3)  & (3.4\%, -1.06, 0.9)  & (4.8\%, -1.13, 10.0)  & (24.8\%, -0.64, 25.5) & (1.0\%, -1.18, 7.9)  & (10.4\%, -1.08, 8.7) & (37.6\%, -0.56, 13.0) \\
& UniRel  & (93.0\%, 0.75, 60.2)  & (70.6\%, 0.15, 3.3)  & (79.0\%, 0.15, 26.8)  & (93.6\%, 0.59, 41.0)  & (72.6\%, 0.05, 15.9)  & (76.4\%, 0.09, 17.7)  & (86.2\%, 0.24, 17.6)  \\ 
\multirow{2}{*}{\textbf{Qwen14B}}        
& Prompting    & (72.0\%, 0.24, 46.9)  & (17.8\%, -0.88, 1.9) & (11.6\%, -0.87, 15.9) & (37.8\%, -0.55, 31.8) & (6.8\%, -1.18, 10.3)  & (16.6\%, -0.84, 11.6) & (41.8\%, -0.41, 13.1) \\
& UniRel  & (97.2\%, 0.83, 63.2)  & ({\color{orange}\textbf{83.6\%}}, {\color{orange}\textbf{0.28}}, {\color{orange}\textbf{3.9}})  & ({\color{orange}\textbf{87.4\%}}, {\color{orange}\textbf{0.22}}, {\color{orange}\textbf{28.1}}) & (98.4\%, 0.67, {\color{orange}\textbf{43.8}})  & (82.8\%, 0.14, 16.7)  & (89.8\%, 0.22, 17.9) & (97.6\%, 0.36, 16.7) \\ 
\midrule
\multirow{2}{*}{\textbf{Llama3B}}         
& Prompting    & (0.0\%, -2.97, 0.0)  & (0.0\%, -2.97, 0.0)  & (0.0\%, -2.92, 0.0)  & (0.0\%, -3.00, 0.0)  & (0.0\%, -2.92, 0.0)  & (0.0\%, -2.93, 0.0)  & (0.0\%, -2.92, 0.0)  \\
& UniRel  & (95.8\%, 0.80, 60.8)  & (70.6\%, 0.14, 2.9)  & (70.6\%, 0.12, 26.7)  & (60.6\%, 0.04, 26.7)  & (38.8\%, -0.47, 12.8) & (75.4\%, 0.09, 14.3)  & (34.6\%, 0.36, 15.4)  \\ 
\multirow{2}{*}{\textbf{Llama8B}}         
& Prompting    & (1.0\%, -2.89, 0.0)  & (0.4\%, -2.74, 0.0)  & (0.0\%, -2.80, 0.0)  & (0.0\%, -2.96, 0.0)  & (0.0\%, -2.67, 0.0)  & (0.2\%, -2.51, 0.0)  & (0.0\%, -2.87, 0.0)  \\
& UniRel  & ({\color{orange}\textbf{98.6\%}}, {\color{orange}\textbf{0.85}}, {\color{orange}\textbf{65.2}})  & (82.6\%, 0.27, 3.9)  & (82.6\%, 0.20, 27.0)  & ({\color{orange}\textbf{99.0\%}}, {\color{orange}\textbf{0.67}}, 42.6)  & ({\color{orange}\textbf{85.0\%}}, {\color{orange}\textbf{0.14}}, {\color{orange}\textbf{17.9}}) & ({\color{orange}\textbf{93.6\%}}, {\color{orange}\textbf{0.25}}, {\color{orange}\textbf{18.9}})  & ({\color{orange}\textbf{98.8\%}}, {\color{orange}\textbf{0.38}}, {\color{orange}\textbf{17.8}})  \\
\midrule
\rowcolor{blue!10}
\multicolumn{2}{l}{\textbf{Optimal Reward}} & 0.92   & 0.77   & 0.65   & 0.72   & 0.65   & 0.67 & 0.79 \\
\bottomrule
\end{tabular}
}
\label{tab:main_results}
\end{table*}

Table~\ref{tab:main_results} presents the performance of Prompting and UniRel across datasets.
Overall, UniRel consistently surpasses Prompting across all evaluated settings.
Across datasets, it delivers at least a \textbf{35\%} improvement in connectivity, a \textbf{245\%} increase in average reward, and a \textbf{27.48\%} gain in Subgraph F1, underscoring its effectiveness in generating valid and informative subgraphs.
Notably, these improvements are observed consistently across datasets with varying structural complexity, indicating that UniRel generalizes well beyond specific graph characteristics.
Within both Qwen and Llama families, larger models demonstrate stronger performance under Prompting, reflecting the benefits of increased model capacity.
UniRel further amplifies these gains, highlighting the complementary benefits of model scaling and task adaptation.

In terms of model-specific results, Qwen-14B achieves the best outcomes on FB15k-237 and MetaQA, while Llama-8B performs best on the remaining datasets. 
On Freebase13 and DBpedia50, performance nearly matches the optimal reward, likely due to the relatively high connectivity of these graphs. 
Reward details are provided in Appendix~\ref{sec:main_results_details}.

\begin{figure*}[t]
    \centering
    \vspace{-0.6em}
    \includegraphics[width=0.9\textwidth]{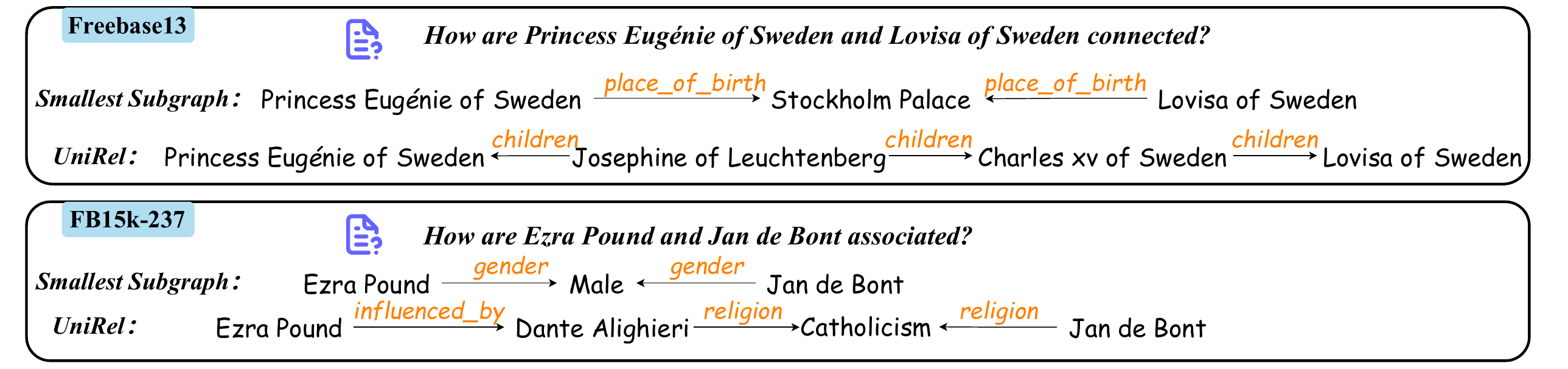} 
    \vspace{-0.3em}
    \caption{Case studies in Freebase13 and FB15k-237.}
    \label{fig:case_studies}
\vspace{-1em}
\end{figure*}

% ~\jh{I suggest using UniRel-R1 (Ours) instead of FT. The word ``Fine-tuned'' sounds like a trivial method but the power of your UniRel-R1 lies  in subgraph pruning, training data curation, and so on, and is much more than fine-tuning. Relevant text descriptions also use the word ``fine-tuned'', and you might want to change them as well. For the baseline method, you can call it ``Prompting''. If you use Zero-shot, maybe reviewers will ask why there isn't a few-shot baseline.}
% ~\jh{Another suggestion: add (Connectivity Ratio, Average Reward) somewhere in the table, as the evaluation metric is only mentioned once and far away.}

\textbf{Case Study.} 
Beyond the quantitative results in Table~\ref{tab:main_results}, we present case studies on multiple datasets to qualitatively assess the outputs of \textit{UniRel}. 
Representative examples are shown in Figure~\ref{fig:case_studies}, where our generated subgraphs are compared with those from the smallest subgraph baseline (i.e.,~with the fewest edges).  
The baseline often relies on trivial relations (e.g., \textit{gender} or \textit{place of birth}), whereas \textit{UniRel} highlights semantically richer intermediates (e.g., \textit{influenced by Dante Alighieri}) that yield more informative explanations.  
This improvement results from pruning, which filters generic nodes, together with RL optimization that promotes unique and meaningful structures.  
Additional case studies are provided in Appendix~\ref{sec:main_results_details}.

\subsection{Generalization} 
\begin{table*}[t]
\centering
\caption{Cross-dataset generalization of UniRel trained on DBpedia500, reported as (connectivity ratio, average reward, subgraph F1).}
\resizebox{0.95\textwidth}{!}{%
\begin{tabular}{llccccc}
\toprule
\multicolumn{2}{c}{\diagbox{\textbf{Datasets}}{\textbf{Models}}} & \textbf{Qwen3B} & \textbf{Qwen7B} & \textbf{Qwen14B} & \textbf{Llama3B} & \textbf{Llama8B} \\
\midrule
\multirow{2}{*}{\textbf{Freebase13}} 
 & \cellcolor{gray!12}Original & (80.0\%, 0.52, 44.7) & (83.4\%, 0.59, 46.1) & (94.0\%, 0.77, 52.7) & (81.8\%, 0.55, 46.9) & (91.0\%, 0.72, 54.1) \\
 & \cellcolor{yellow!12}Modified & (69.6\%, 0.33, 41.3) & (74.8\%, 0.43, 44.2) & (84.0\%, 0.60, 50.4) & (81.6\%, 0.55, 46.9) & (87.0\%, 0.65, 51.2) \\
\midrule
\multirow{2}{*}{\textbf{MetaQA}}     
 & \cellcolor{gray!12}Original & (60.6\%, -0.12, 20.0) & (69.8\%, 0.04, 21.7) & (78.0\%, 0.11, 23.1) & (64.2\%, -0.06, 21.7) & (80.0\%, 0.18, 23.9) \\
 & \cellcolor{yellow!12}Modified & (29.2\%, -0.56, 9.1)  & (49.0\%, -0.21, 16.5)  & (60.8\%, -0.02, 18.4)  & (67.2\%, 0.06, 22.5) & (72.6\%, 0.15, 23.2) \\
\midrule
\multirow{2}{*}{\textbf{YAGO3-10}}   
 & \cellcolor{gray!12}Original & (56.6\%, -0.18, 11.5) & (67.4\%, 0.01, 13.2) & (80.6\%, 0.15, 15.0) & (44.8\%, -0.36, 10.1) & (83.0\%, 0.18, 15.8) \\
 & \cellcolor{yellow!12}Modified & (23.6\%, -0.66, 7.23) & (41.4\%, -0.34, 10.0) & (48.2\%, -0.24, 11.4) & (44.6\%, -0.31, 10.5) & (62.2\%, -0.03, 13.8) \\
\bottomrule
\end{tabular}
}
\label{tab:generalization_results}
\end{table*}

\begin{figure*}[t]
    \centering
    \begin{subfigure}{0.33\textwidth}
        \centering
        \includegraphics[width=\linewidth]{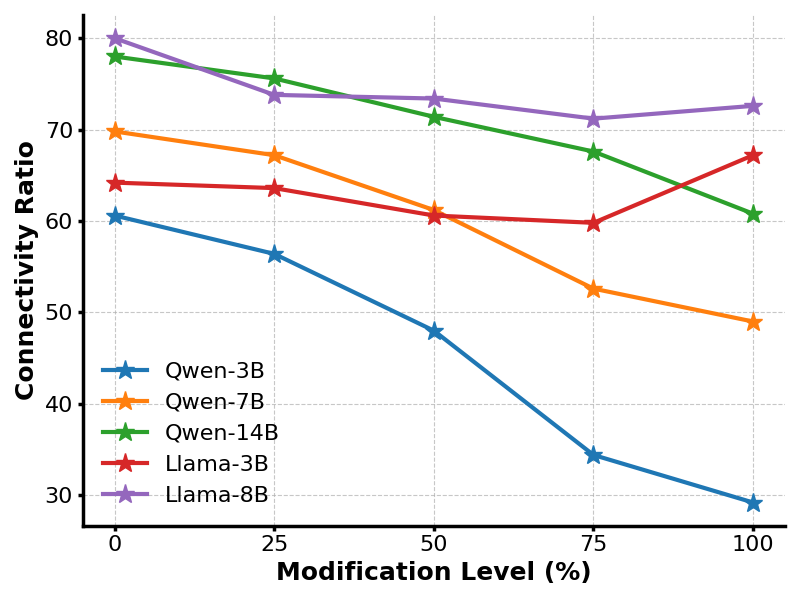}
        \vspace{-0.5em}
        %\caption{Connectivity Ratio v.s. Modification Level}
        \label{fig:c_vs_mod}
    \end{subfigure}
    \hfill
    \begin{subfigure}{0.33\textwidth}
        \centering
        \includegraphics[width=\linewidth]{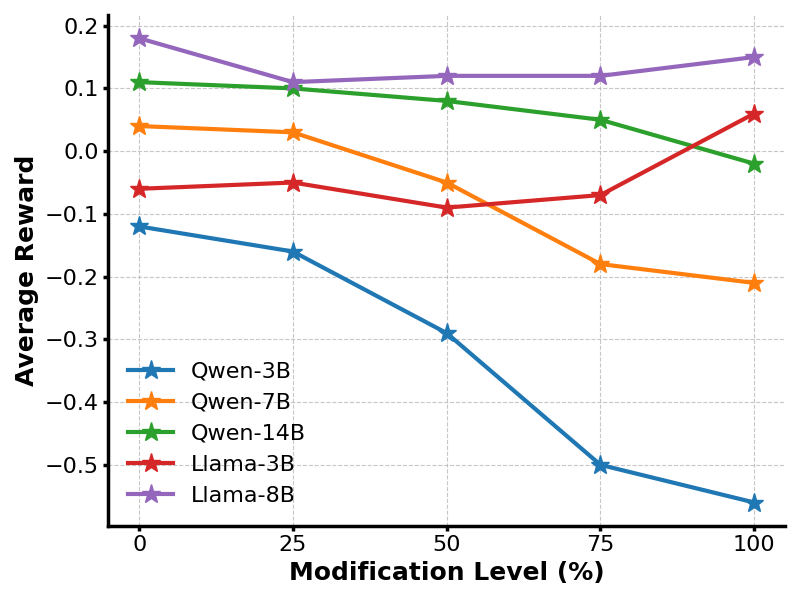}
        \vspace{-0.5em}
        %\caption{Average Reward v.s. Modification Level}
        \label{fig:f1_vs_mod}
    \end{subfigure}
    \hfill
    \begin{subfigure}{0.33\textwidth}
        \centering
        \includegraphics[width=\linewidth]{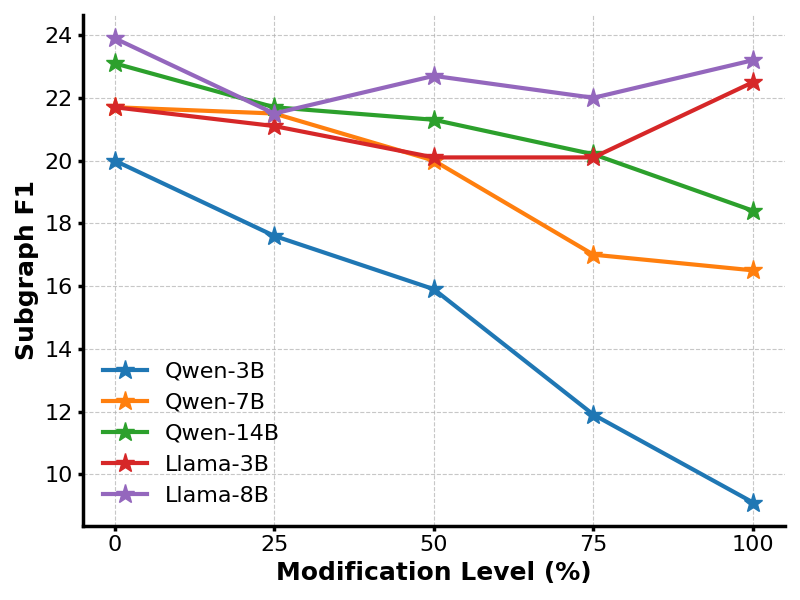}
        \vspace{-0.5em}
        %\caption{Subgraph F1 v.s. Modification Level}
        \label{fig:r_vs_mod}
    \end{subfigure}
    \vspace{-2em}
    \caption{Comparison of connectivity ratio, average reward, and subgraph F1 under different modification levels in MetaQA.}
    \label{fig:comparison_generalization}
    \vspace{-0.8em}
\end{figure*}

To examine the robustness and transferability of \textit{UniRel}, we evaluate its ability to generalize across datasets.
Table~\ref{tab:generalization_results} reports partial generalization results, with the complete results provided in Appendix~\ref{sec:generalization_results_details}.
Specifically, UniRel is trained on the largest dataset, DBpedia500, and directly evaluated on the remaining datasets without additional training.

\textbf{Settings.} \textbf{\emph{Original}} denotes the standard setting, where datasets retain their original entity and relation names.
\textbf{\emph{Modified}} denotes a controlled variant in which all entities and relations are replaced with random identifiers (e.g., $\text{ENT}1,\dots,\text{ENT}n$, $\text{REL}1,\dots,\text{REL}m$), thereby removing semantic cues while preserving graph structure.

\textbf{Result Analysis.} As shown in the \emph{Original} rows, the models are able to generate valid subgraphs even for previously unseen entities and relations, demonstrating a notable degree of cross-domain transferability.
Compared with the Prompting results in Table~\ref{tab:main_results}, these models achieve substantially higher performance across all metrics, often approaching the connectivity, average reward, and Subgraph F1 of their in-domain UniRel counterparts, suggesting that UniRel generalizes effectively beyond the training domain.

% This observation motivates the hypothesis that LLMs exploit semantic regularities acquired during training to generalize to novel patterns.
% To test this hypothesis, we evaluate UniRel under the \emph{Modified} setting.

This motivates the hypothesis that LLMs leverage semantic regularities to generalize.
We test this by evaluating UniRel under the \emph{Modified} setting.
% This observation motivates the hypothesis that LLMs can exploit semantic regularities acquired during training to generalize to novel patterns.  
% To test this, we construct \emph{Modified} versions of each dataset by replacing all entities and relations with random identifiers (e.g., $\text{ENT}1,\dots,\text{ENT}n$, $\text{REL}1,\dots,\text{REL}m$), thereby eliminating semantic cues.

As expected, the \emph{Modified} rows exhibit a substantial performance drop across all datasets for the Qwen family, highlighting the critical role of semantic information in enabling cross-domain transfer.
In contrast, the Llama family is less affected: the 3B model shows only marginal changes, and although the 8B model does experience performance degradation, the magnitude remains considerably smaller than that observed for Qwen models.
For example, on YAGO3-10, Qwen-3B suffers reductions of \textbf{58.3\%} in connectivity ratio, \textbf{266.67\%} in average reward, and \textbf{37.13\%} in Subgraph F1, whereas Llama-3B shows much smaller changes of \textbf{0.45\%}, \textbf{13.89\%}, and \textbf{3.96\%}, respectively.
Overall, these results indicate that Qwen models are more sensitive to the removal of semantic information than Llama models.

To further probe this effect, we conduct an additional experiment on MetaQA, where a fraction of entities and relations are replaced with random identifiers.
We consider partial modifications at 25\%, 50\%, and 75\%, thereby progressively diminishing the amount of preserved semantic information.
As shown in Figure~\ref{fig:comparison_generalization}, Qwen models exhibit a consistent degradation in connectivity ratio, average reward, and Subgraph F1 as the modification level increases, indicating that both feasibility and structural quality of the generated subgraphs strongly depend on semantic regularities in the knowledge graph.
By contrast, Llama models are less affected: the 3B variant shows only minor fluctuations, with a slight initial decline followed by a modest recovery, whereas the 8B model exhibits a mild downward trend, with reductions of \textbf{9.25\%} in connectivity ratio, \textbf{16.67\%} in average reward, and \textbf{2.93\%} in Subgraph F1 between the original and fully modified settings.

Taken together, these findings suggest that Qwen models rely more on semantic information for cross-dataset generalization, whereas Llama models draw more on structural connectivity to sustain stable performance.

\vspace{-0.5em}
\subsection{Scalability}
\begin{table*}[t]
\centering
\caption{Performance comparison of Prompting and UniRel on DBpedia50 across three-entity and four-entity queries.}
\resizebox{0.9\textwidth}{!}{%
\begin{tabular}{ll|cccc|ccccc}
\toprule
\multicolumn{2}{c|}{\textbf{Models}}
 & \multicolumn{4}{c|}{\textbf{Three Entities}} 
 & \multicolumn{5}{c}{\textbf{Four Entities}} \\ 
\multicolumn{2}{l|}{} 
 & \makecell{Full Conn. } 
 & \makecell{Pairwise Conn.} 
 & {Reward} 
 & {F1} 
 & \makecell{Full Conn.} 
 & \makecell{Triple Conn.} 
 & \makecell{Pairwise Conn.} 
 & {Reward}
 & {F1} \\ 
\midrule 
\multirow{2}{*}{Qwen3B} 
 & Prompting    & 4.2\% & 2.2\% & -1.21 & 6.9 & 1.8\% & 0.0\% & 0.6\% & -2.31 & 2.5 \\
 & UniRel  & 69.2\% & 17.2\% & 1.17 & 44.6 & 8.2\% & 18.6\% & 50.8\% & -0.29 & 35.2\\

\multirow{2}{*}{Qwen7B} 
 & Prompting    & 10.4\% & 6.6\% & -0.66 & 18.6 & 5.2\% & 2.0\% & 4.8\% & -0.84 & 12.1\\
 & UniRel  & 73.4\% & 18.0\% & 1.27 & 45.5 & 55.4\% & 16.4\% & 18.0\% & 0.78 & 43.7\\

\multirow{2}{*}{Qwen14B} 
 & Prompting    & 13.8\% & 13.6\% & -0.56 & 29.3 & 4.8\% & 0.6\% & 5.6\% & -2.37 & 22.0 \\
 & UniRel  & 95.4\% & 2.0\%  & 1.59 & 50.0 & 83.2\% & 6.1\% & 4.3\% & 1.26 &  46.9\\ 
\midrule

\multirow{2}{*}{Llama3B} 
 & Prompting    & 0.0\% & 0.0\% & -2.93 & 0.0   & 0.0\% & 0.0\% & 0.0\% & -3.82 & 0.0\\
 & UniRel  & 90.8\% & 7.6\% & 1.57 & 50.2  & 48.4\% & 18.4\% & 24.2\% & 0.68 & 43.9\\

\multirow{2}{*}{Llama8B} 
 & Prompting    & 0.0\% & 0.0\% & -2.90 & 0.0 & 0.0\% & 0.0\% & 0.0\% & -3.78 & 0.0 \\
 & UniRel  & 94.6\% & 4.8\% & 1.61 & 51.1 & 80.0\% & 5.6\% & 10.2\% & 1.26 & 46.8\\

\midrule
\rowcolor{blue!10}
\multicolumn{2}{l|}{\textbf{Optimal Reward}} & \multicolumn{4}{c|}{1.74} & \multicolumn{5}{c}{1.63} \\
\bottomrule
\end{tabular}}
\vspace{-1em}
\label{tab:multi_entity_results}
\end{table*}

To assess the scalability of \textit{UniRel}, we extend our evaluation from two-entity queries to queries involving three and four entities on the DBpedia50 dataset.
Unlike the two-entity case, where connectivity is measured by a \emph{connectivity ratio}, multi-entity queries allow for a richer characterization.

For three-entity queries, we report both the \emph{pairwise connectivity ratio} (the percentage of queries where \textit{only} two entities are connected) and the \emph{full connectivity ratio} (all three entities are connected).
For four-entity queries, we distinguish three levels: (i) \emph{pairwise connectivity ratio}, (ii) \emph{triple connectivity ratio}, and (iii) \emph{full connectivity ratio}.

Table~\ref{tab:multi_entity_results} summarizes the performance of Prompting and UniRel on DBpedia50 across three-entity and four-entity queries.
Across both settings, Prompting models exhibit limited ability to recover valid subgraphs.
By contrast, UniRel demonstrates substantial generalization gains: for three-entity queries, the full connectivity ratio increases by \textbf{591.3\%}, the average reward by \textbf{277.27\%}, and the Subgraph F1 by \textbf{70.65\%};
for four-entity queries, the corresponding gains reach \textbf{355.6\%} in full connectivity, \textbf{87.5\%} in reward, and \textbf{113.1\%} in Subgraph F1.
Importantly, even when full connectivity is not achieved, UniRel consistently improves pairwise and triple connectivity ratios together with Subgraph F1, indicating its ability to recover meaningful partial structures that explain subsets of entity relations.
Moreover, Llama models consistently outperform Qwen models, in some cases approaching the optimal reward, highlighting their stronger capacity to leverage structural information in multi-entity reasoning.

These findings collectively demonstrate the scalability of the proposed framework to more complex multi-entity scenarios and underscore its applicability to real-world KGQA tasks that involve reasoning over multiple entities.

% \begin{table}[h]
% \centering
% \caption{Performance Comparison with RL-tuned LLM}
% \begin{tabular}{lcccc}
% \toprule
% \textbf{Model} & \textbf{Configuration} & \textbf{Hit@1} & \textbf{F1} \\
% \midrule
% \multirow{2}{*}{G-retriever} & Base & 70.49 & 53.41 \\
%  & Retriever + RL-tuned & 70.10 & \textbf{58.90} \\
% \midrule
% \multirow{2}{*}{ROG} & Base & \textbf{85.70} & 70.80 \\
%  & Retriever + RL-tuned & 79.00 & \textbf{72.30} \\
% \midrule
% \multirow{2}{*}{KAPING} & Base & 52.64 & - \\
%  & Retriever + RL-tuned & 71.40 & 61.70 \\
% \midrule
% \multirow{2}{*}{GNN} & Base & 80.60 & \textbf{71.30} \\
%  & Retriever + RL-tuned & 80.40 & 70.90 \\
% \midrule
% \multirow{2}{*}{GNN-RA} & Base & 82.80 & 73.50 \\
%  & Retriever + RL-tuned & 82.50 & \textbf{77.00} \\
% \bottomrule
% \end{tabular}
% \end{table}

\begin{table}[t]
\centering
\caption{Comparison results for entity-centric KGQA.}
\resizebox{0.5\textwidth}{!}{%
\begin{tabular}{llcccc}
\toprule
\multirow{2}{*}{Retriever} & \multirow{2}{*}{Pipeline}
& \multicolumn{2}{c}{WebQSP} & \multicolumn{2}{c}{CWQ} \\
\cmidrule(lr){3-4} \cmidrule(lr){5-6}
& & Hit@1 & F1 & Hit@1 & F1 \\
\midrule
PCST (G-retriever) & Original & 70.49 & 53.41 & -- & -- \\
~\cite{he2024g}    & UniRel   & \textbf{70.50} & \textbf{58.90} & 49.53 & 44.70 \\
\midrule
KAPING Retriever   & Original & 52.64 & -- & -- & -- \\
~\cite{baek2023knowledge}  & UniRel   & \textbf{72.00} & 61.40 & 50.40 & 46.20 \\
\midrule
ROG Retriever      & Original & \textbf{85.70} & 70.80 & \textbf{62.60} & 56.20 \\
~\cite{luo2024reasoning}& UniRel   & 80.00 & \textbf{72.30} & 59.2 & \textbf{57.20} \\
\midrule
GNN Retriever      & Original & \textbf{80.60} & \textbf{71.30} & 61.70 & 59.40 \\
~\cite{mavromatis2024gnn} & UniRel   & 80.20 & 71.10 & \textbf{65.20} & \textbf{61.70} \\
\midrule
GNN-RA Retriever   & Original & \textbf{82.80} & 73.50 & 62.80 & 62.40 \\
~\cite{mavromatis2024gnn} & UniRel   & 82.70 & \textbf{77.10} & \textbf{66.50} & \textbf{63.70} \\
\bottomrule
\end{tabular}
}
\label{tab:entity_centric}
\vspace{-1em}
\end{table}

\subsection{UniRel for Entity-Centric KGQA}
Although UniRel is designed for relation-centric KGQA, its pipeline is task-agnostic and can be readily adapted to conventional entity-centric KGQA.
Specifically, we replace the subgraph retriever module with task-specific entity-centric retrievers, while keeping the remaining reasoning and generation components unchanged.

We instantiate this adaptation using retrievers from prior work, and all models use LLaMA2-7B-chat~\cite{touvron2023llama} as the base LLM.
Evaluation is conducted on  WebQSP~\cite{yih2016value} and CWQ~\cite{talmor2018web} using standard Hit@1 and F1 metrics. More details are provided in Appendix~\ref{sec:entity_centric_results_details}.

% As shown in Table~\ref{tab:entity_centric}, under the same retriever, replacing the original reasoning component with the UniRel pipeline yields competitive and often improved performance, demonstrating that UniRel can be effectively applied to entity-centric KGQA.

% Table~\ref{tab:entity_centric} reports the results of applying the UniRel pipeline to entity-centric KGQA under different retrievers.
% For each retriever, we compare UniRel with the original pipeline from prior work under the same retriever and base LLM, showing consistent F1 improvements and competitive or improved Hit@1 scores.
% Table~\ref{tab:entity_centric} reports results of applying UniRel to entity-centric KGQA under different retrievers.
% For each retriever, we compare UniRel with the original pipeline reported in prior work under the same retriever and base LLM, showing consistent F1 improvements and competitive or improved Hit@1 scores.
Table~\ref{tab:entity_centric} reports the results of applying the UniRel pipeline to entity-centric KGQA under different retrievers. For each retriever, we compare the original pipeline with UniRel while keeping the retriever and base LLM fixed. The results of the original pipelines are taken directly from the prior work. Overall, UniRel consistently improves F1 across most settings and achieves competitive or improved Hit@1 scores.
Notably, under PCST and KAPING retrievers, UniRel substantially improves F1 on WebQSP, indicating more complete answer coverage.
For stronger retrievers such as GNN and GNN-RA, UniRel yields consistent gains on both WebQSP and CWQ, suggesting that improved relational reasoning can further benefit entity-centric QA when high-quality subgraphs are available.

\section{Related Work}

\textbf{Knowledge Graph Question Answering.}
Recent progress in KGQA has incorporated LLMs to exploit their strong natural language understanding. A prominent direction is to generate direct answers by enriching LLMs with KG evidence, most commonly via RAG frameworks that retrieve relevant triples as context~\citep{linders2025knowledge}. Another line of work integrates LLMs with GNNs for joint reasoning over text and structured knowledge~\citep{xu2025harnessing,yasunaga2021qa,he2024g}. In parallel, research emphasizing explainability focuses on explicit reasoning paths for multi-hop questions, where the answer requires traversing multiple triples~\citep{zhou2018interpretable,chakraborty2024multi,zhang2018variational}. Such methods often employ RL or search strategies, including MCTS, to iteratively refine candidate paths~\citep{shen2025reasoning}. Others adopt embedding-based reasoning approaches~\citep{saxena2020improving,shi2021transfernet}, where multi-hop inference is captured implicitly through learned representations across layers.

\textbf{Improving LLM Reasoning via RLVR.}
Recent work has demonstrated RLVR's broad applicability across a spectrum of tasks, including mathematical and logical reasoning~\citep{guo2025deepseek}, code generation~\citep{wang2025code}, multi-modal reasoning~\citep{huang2025vision, wang2025vl,li2025self}, structured data tasks like relation extraction~\citep{dai2025r1} and GUI navigation~\citep{shi2025mobilegui}, and complex reasoning strategies such as parallel thinking~\citep{zheng2025parallel}. Concurrent research focuses on improving the framework through two main directions: exploring training paradigms like self-play~\citep{liu2025spiral, huang2025r} and test-time reinforcement learning~\citep{zuo2025ttrl}, and developing more effective RL algorithms such as DAPO~\citep{yu2025dapo}, VAPO~\citep{yue2025vapo}, and high-entropy guided optimization techniques~\citep{dai2025cde,wang2025beyond,Zhou2025EvolvingLM}.

\section{Conclusion}

In this paper, we introduced relation-centric KGQA, where the goal is to return a subgraph that captures semantic associations among entities, a complementary setting to the standard entity-centric KGQA. 
We presented \textit{UniRel}, a unified modular framework that integrates subgraph retriever with RL-tuned LLMs.
By design, UniRel favors compact and specific subgraphs through a reward function that emphasizes informative relations and low-degree intermediate entities, steering the model toward concise yet distinctive relational answers.
Extensive experiments demonstrate that UniRel consistently improves connectivity and reward over Prompting baselines and generalizes effectively to unseen entities and relations.
Our analysis further revealed a model-level distinction: Qwen models exhibit a stronger reliance on semantic cues, whereas Llama models maintain more stable performance by leveraging structural connectivity.
Also, our results on multi-entity queries confirm the scalability of UniRel, showing that the framework extends naturally to complex reasoning tasks.
Beyond the relation-centric setting, we further showed that the UniRel pipeline can be applied to conventional entity-centric KGQA by replacing the subgraph retriever module, achieving competitive or improved Hit@1 and F1 scores across multiple benchmarks.

% In this paper, we introduced relation-centric KGQA, where the goal is to return a subgraph that captures semantic associations among entities, a complementary setting to the standard entity-centric KGQA. 
% We presented \textit{UniRel-R1}, a unified framework that integrates subgraph selection, multi-stage pruning, and RL-tuned LLMs, where the reward design explicitly favors compact subgraphs with rarer relations and lower-degree intermediates, thereby steering the model toward concise and distinctive relational answers.
% Through extensive evaluations on seven benchmark knowledge graphs, we demonstrated that UniRel-R1 delivers substantial improvements over Prompting baselines, achieving large gains in both connectivity and reward while maintaining strong generalization to unseen entities and relations.
% Our analysis further revealed a model-level distinction: Qwen models exhibit a stronger reliance on semantic cues, whereas Llama models maintain more stable performance by leveraging structural connectivity.
% Finally, our results on multi-entity queries confirm the scalability of UniRel-R1, showing that the framework extends naturally to complex reasoning tasks.

\section*{Impact Statements}
This work studies relation-centric knowledge graph question answering, where the goal is to generate structured subgraphs that explicitly describe how entities are related. By combining principled subgraph retrieval with RL–tuned LLMs, UniRel promotes compact, informative, and interpretable relational explanations. This setting may support Human-AI collaboration in applications that require transparent reasoning over structured knowledge, such as education, scientific exploration, and decision support.

Potential risks arise from misuse or over-reliance on automatically generated relational explanations. The quality and neutrality of the produced subgraphs depend on the underlying knowledge graph and reward design, which may reflect incompleteness or bias. As a result, generated explanations could be misinterpreted as comprehensive or authoritative in sensitive contexts.

Overall, this work aims to encourage more interpretable and controllable knowledge-based reasoning to better align AI systems with human understanding. To mitigate potential risks, future work may explore uncertainty modeling, user preferences, and forms of human feedback to support responsible and collaborative deployment.

\bibliography{example_paper}
\bibliographystyle{icml2026}

\newpage
\appendix
\onecolumn
% \section*{Reproducibility Statement}
% We ensure reproducibility of our work as follows. The prompts used in all experiments are provided in Appendix~\ref{sec:implement_details}, and a complete list of hyperparameters is included in Appendix~\ref{sec:hyperparameters}. All datasets and models employed in our experiments are publicly available. In addition, detailed descriptions of dataset modifications and query constructions are documented in Appendix~\ref{sec:implement_details}.

% \section*{The Use of Large Language Models}
% This manuscript was refined with the assistance of LLMs, which were used to improve clarity, grammar, and overall readability. The use of these models was restricted to language editing and did not influence the scientific content.
\section{Complexity Analysis of the Subgraph Retriever}\label{sec:complexity}
\subsection{Subgraph Selection}
Computing the $k$-hop neighborhood for a single seed $e_i$ can be implemented via a truncated BFS on the undirected projection of the knowledge graph, which takes 
$O(|\mathcal V^k(e_i)| + |\mathcal T^k(e_i)|)$ time.
Aggregating over all $m$ seed entities, the total time complexity of subgraph selection is
\[
O\!\left(\sum_{i=1}^{m} \big(|\mathcal V^k(e_i)| + |\mathcal T^k(e_i)|\big)\right).
\]
Maintaining the union of nodes and edges using hash-based data structures incurs only linear overhead in the size of the resulting candidate graph and does not change the asymptotic complexity.

\subsection{Multi-step Graph Pruning}
Computing $\deg(v)$ for all $v\in\mathcal V'$ can be done by a single pass over edges in $\mathcal T'$, which takes $O(|\mathcal T'|)$ time.

\paragraph{Stage 1: Local Pruning.}
Thresholding nodes by $\rho$ requires scanning all nodes once, i.e., $O(|\mathcal V'|)$.
Removing incident edges can be implemented by filtering $\mathcal T'$ in one pass (keeping only edges whose endpoints survive), which costs $O(|\mathcal T'|)$.
Finally, discarding isolated nodes can be done by recomputing degrees on the remaining edge set, adding another $O(|\mathcal T'|)$ (or $O(|\mathcal V'|+|\mathcal T'|)$ with an adjacency list).
Overall, the time complexity of Stage~1 is
\[
O(|\mathcal V'| + |\mathcal T'|).
\]

Note that $\rho$ only affects the size of the retained node/edge sets after pruning; the worst-case running time remains linear in $|\mathcal V'|+|\mathcal T'|$.

\paragraph{Stage 2: Connectivity Guarantee.}
Assume each pruned neighborhood is stored as a hash set.
Then computing one intersection $\mathcal I_{ij}$ takes
$O(\min\{|\mathcal V^k_\rho(e_i)|, |\mathcal V^k_\rho(e_j)|\})$ time by iterating over the smaller set and checking membership in the larger set.
Thus, computing all pairwise intersections costs
\[
O\!\left(\sum_{1\le i<j\le m}\min\Big\{|\mathcal V^k_\rho(e_i)|, |\mathcal V^k_\rho(e_j)|\Big\}\right).
\]
% Let $N_k(\rho)=\max_i |\mathcal V^k_\rho(e_i)|$. 
In the worst case, this is upper bounded by
$
O(m^2\,\max_i |\mathcal V^k_\rho(e_i)|).
$
Building $\mathcal H_\rho$ while checking whether $\mathcal I_{ij}$ is empty is included in the above cost, and checking connectivity of $\mathcal H_\rho$ via BFS/DFS takes
$O(m+|\mathcal T_\rho|)\le O(m^2)$,
where $T_\rho$ is the edges of $\mathcal H_\rho$.
Let
$
N_k(\rho)=\max_i |\mathcal V^k_\rho(e_i)|
$. Hence, the overall per-round time complexity of Stage~2 is dominated by
\[
O(m^2 N_k(\rho)).
\]

Let $R(\rho)$ be the number of relaxation rounds until $\mathcal H_\rho$ becomes connected.
Since each round reruns Stages~1--2, the total time spent on Stage~2 across all rounds is
\[
O\!\left(\sum_{t=1}^{R(\rho)} m^2\,N_k(\rho_t)\right)
\quad\le\quad
O\!\left(m^2 R(\rho)  N_k(\rho^\star))\right),
\]
where $\rho^\star$ denotes the final threshold at which connectivity is restored.

\paragraph{Stage 3: Leaf Entity Pruning.}
This process can be implemented using a queue-based peeling procedure.
We first compute the degree of each node in $O(|\mathcal T_\rho|)$ time, enqueue all non-seed nodes with degree one, i.e., with only one neighbor. and then iteratively remove them while updating the degrees of their neighbors.
Each node is enqueued and removed at most once, and each relation in $\mathcal T_\rho$ is deleted at most once.
Therefore, the total time complexity of Stage~3 is
\[
O\!\left(|\mathcal V_\rho| + |\mathcal T_\rho|\right).
\]

\paragraph{Stage 4: Compactness Control.}
Let $\mathcal I^+=\{\mathcal I_{ij}\mid \mathcal I_{ij}\neq\emptyset\}$ denote the collection of non-empty intersections obtained in Stage~2,
where $\mathcal I_{ij}=\mathcal V^k_\rho(e_i)\cap\mathcal V^k_\rho(e_j)$.
We select a subset $T^c\subseteq \mathcal I^+$ whose index pairs jointly cover all seed entities.
Let $\lambda=|T^c|$ denote the number of selected intersections.

For each $\mathcal I_{ij}\in T^c$, we retain a subset $\hat{\mathcal I}_{ij}\subseteq\mathcal I_{ij}$ of size $s$
corresponding to the entities with the lowest hub penalties.
This operation can be implemented by partial selection (e.g., using a heap of size $s$),
which takes $O(|\mathcal I_{ij}|\log s)$ time. Since 
$
N_k(\rho)=\max_i |\mathcal V^k_\rho(e_i)|$
and $|\mathcal I_{ij}|\le N_k(\rho)$, the total cost of this step is
\[
O\!\left(\sum_{\mathcal I_{ij}\in T^c} |\mathcal I_{ij}|\log s\right)
\le O\big(\lambda\,N_k(\rho)\log s\big).
\]

Next, for each retained set $\hat{\mathcal I}_{ij}$, we perform a truncated $k$-hop expansion restricted to the corresponding
pruned neighborhoods $\mathcal V^k_\rho(e_i)\cup\mathcal V^k_\rho(e_j)$.
This expansion can be implemented as a multi-source BFS and visits each node and relation at most once within the restricted region.
Let
$
T(\rho)=\max_i |\mathcal T^k_\rho(e_i)|
$
denote the maximum number of relations in a pruned $k$-hop neighborhood.
The cost of this expansion is $O(N_k(\rho)+T_k(\rho))$ per intersection,
and $O(\lambda(N_k(\rho)+T_k(\rho)))$ in total.

For each $T^c$, the induced subgraph over the expanded node set is constructed and further simplified using Stage~3.
Both operations are linear in the size of the resulting subgraph and thus do not affect the asymptotic complexity.

Putting everything together, the overall time complexity of Stage~4 is
\[
O\big(\lambda\,N_k(\rho)\log s + \lambda(N_k(\rho)+T_k(\rho))\big).
\]

\paragraph{Stage 5: Iterative Reduction.}
Let $R (s)$ denote the number of reduction rounds, i.e., the number of times Stage~4 is reapplied with decreasing values of $s$.
Since each round invokes Stage~4 once, the total cost of Stage~5 is the sum of the costs across all rounds.
Using the complexity of Stage~4, the overall time complexity of Stage~5 is
\[
O\!\left(
\sum_{t=1}^{R (s)}
\big(
\lambda\,N_k(\rho)\log s_t
+
\lambda(N_k(\rho)+T_k(\rho))
\big)
\right),
\]
where $s_t$ denotes the value of $s$ at the $t$-th reduction round.

In practice, $s$ is initialized to a small constant and monotonically decreased, and $R (s)$ is typically very small.
Therefore, the complexity of Stage~5 is dominated by a constant number of invocations of Stage~4 and remains
\[
O\big(\lambda(N_k(\rho)+T_k(\rho))\big)
\]
up to a small multiplicative factor.

\subsection{Overall Complexity}

We summarize the time complexity of the entire subgraph retrieval and pruning pipeline.
Recall that $m=|\mathcal E_q|$ denotes the number of seed entities.
We denote by
$
N_k(\rho)=\max_i |\mathcal V^k_\rho(e_i)|, \qquad
T_k(\rho)=\max_i |\mathcal T^k_\rho(e_i)|
$
the maximum numbers of nodes and relations in a pruned $k$-hop neighborhood under threshold $\rho$.
Here, $R(\rho)$ denotes the number of threshold relaxation rounds in Stage~2,
$\lambda=|T^c|$ the number of selected intersections in Stage~4,
and $R(s)$ the number of reduction rounds in Stage~5.

Aggregating the costs of all stages, the total time complexity is
\[
T_{\text{total}}
=
O\!\left(
\sum_{i=1}^{m}\big(|\mathcal V^k(e_i)|+|\mathcal T^k(e_i)|\big)
+
m^2\,R(\rho)\,N_k(\rho^*)
+
R(s)\big(\lambda\,N_k(\rho^*)\log s + \lambda(N_k(\rho^*)+T_k(\rho^*))\big)
+
|\mathcal V'_{\rho^*}|+|\mathcal T'_{\rho^*}|
\right),
\]
where $\rho^*$ denotes the final threshold at which connectivity is restored.

We simplify this expression by identifying dominated terms.
First, since $\mathcal V'_{\rho^*}\subseteq \bigcup_{i=1}^m \mathcal V^k_{\rho^*}(e_i)$ and
$\mathcal T'_{\rho^*}\subseteq \bigcup_{i=1}^m \mathcal T^k_{\rho^*}(e_i)$, we have the crude upper bounds
\[
|\mathcal V'_{\rho^*}| \le \sum_{i=1}^m |\mathcal V^k_{\rho^*}(e_i)| \le m\,N_k(\rho^*),
\qquad
|\mathcal T'_{\rho^*}| \le \sum_{i=1}^m |\mathcal T^k_{\rho^*}(e_i)| \le m\,T_k(\rho^*).
\]
Hence the linear pruning terms satisfy
\[
|\mathcal V'_{\rho^*}|+|\mathcal T'_{\rho^*}| = O\big(m\,(N_k(\rho^*)+T_k(\rho^*))\big).
\]
Similarly, the subgraph selection cost can be upper bounded by
\[
\sum_{i=1}^{m}\big(|\mathcal V^k(e_i)|+|\mathcal T^k(e_i)|\big)
= O\big(m\,(N_k(\rho^*)+T_k(\rho^*))\big),
\]
up to a change from $\rho$-pruned neighborhoods to an upper bound.

Second, the term $\lambda\,N_k(\rho^*)\log s$ arises from partial selection within intersections.
In practice $s$ is a small constant, so $\log s = O(1)$ and
\[
\lambda\,N_k(\rho^*)\log s = O(\lambda\,N_k(\rho^*)).
\]

Third, Stage~5 repeats Stage~4 for $R(s)$ rounds with decreasing $s$.
In practice $R(s)$ is typically very small (often a constant), so Stage~5 introduces only a small multiplicative factor.

Combining these observations, and dropping lower-order linear terms that are dominated by the costs in Stage~2 and Stages~4--5,
the overall complexity is dominated by
\[
T_{\text{total}}
=
O\!\left(
m^2\, R(\rho)\,N_k(\rho^*)
+
R(s)\,\lambda\,(N_k(\rho^*)+T_k(\rho^*))
\right).
\]

\subsection{Discussion}
A natural formulation for connecting multiple seed entities is Steiner tree problem.
While Steiner-tree approaches are theoretically appealing, their computational characteristics
and operational assumptions differ substantially from the requirements of query-driven reasoning on large knowledge graphs.
Below we discuss key considerations motivating our design.

\paragraph{Computational complexity.}
Exact Steiner tree computation on general graphs is NP-hard and exhibits exponential dependence on the number of terminals $m$.
Typical exact algorithms, such as dynamic programming over subsets of terminals, have worst-case running time
\[
O\!\left(3^{m}\,|\mathcal V| + 2^{m}\,|\mathcal V|\log|\mathcal V|\right),
\]
which is up to polylogarithmic factors and quickly becomes prohibitive even for moderately sized multi-entity queries.

Approximate or heuristic Steiner methods avoid the exponential dependence on $m$,
but their complexity is typically dominated by repeated shortest-path computations among terminals.
For example, constructing a metric closure by running single-source shortest-path algorithms from each terminal
followed by a minimum spanning tree incurs
\[
O\!\left(m\,(|\mathcal T| + |\mathcal V|\log|\mathcal V|) + m^2\log m\right),
\]
 where the dominant term arises from $m$ shortest-path searches on the (sub)graph.

In contrast, our approach avoids exponential dependence on $m$.
As shown in our analysis, the dominant cost scales as
\[
O\!\left(
R(\rho)\,m^2\,N_k(\rho^*)
+
R(s)\,\lambda\,(N_k(\rho^*)+T_k(\rho^*))
\right),
\]
which is polynomial in $m$ and bounded by the sizes of pruned $k$-hop neighborhoods.

\textbf{Locality of computation.}
A key distinction lies in the scope of computation.
Steiner-based formulations typically rely on shortest-path searches whose cost is tied to global graph structure.
In contrast, all major operations in our framework: pairwise intersections, connectivity checks,
and compactness control—are confined to pruned local neighborhoods
$\mathcal V^k_{\rho^*}(e_i)$ and their unions.
Consequently, the complexity is governed by $N_k(\rho^*)$ and $T_k(\rho^*)$ rather than global graph statistics.

\textbf{Controlled trade-off between connectivity and compactness.}
Our method exposes explicit parameters $(\rho,k,\lambda,s)$ that directly regulate the trade-off
between connectivity and compactness.
The threshold $\rho$ and the number of relaxation rounds $R(\rho)$ control how aggressively neighborhoods are pruned,
while $\lambda$ and $s$ bound the number and size of retained intersections.
This results in a predictable and tunable computational budget at query time.
By contrast, Steiner-tree formulations do not provide similarly explicit query-level controls over the explored subgraph size.

\textbf{Pairwise interactions versus global optimization.}
Although our approach involves a quadratic dependence on $m$ due to pairwise interactions among seed entities,
these interactions are limited to set intersections and bounded neighborhood expansions.
This is fundamentally different from global optimization objectives in Steiner-tree formulations,
where pairwise relationships are mediated through shortest-path computations over large graphs.
As a result, the quadratic term in our method remains tractable in practice and does not introduce super-polynomial behavior.

\section{Details of Experiments}

\subsection{Hyperparameters}\label{sec:hyperparameters}
Table~\ref{tab:hyperparameters} summarizes the key hyperparameters used in our experiments.

\begin{table}[h]
\centering
\caption{Hyperparameter settings used across all experiments.}
\resizebox{0.5\textwidth}{!}{%
\begin{tabular}{ll}
\toprule
\textbf{Category} & \textbf{Setting} \\
\midrule
\multicolumn{2}{l}{\textbf{Data / Input Configuration}} \\
Max Prompt Length        & 8192 \\
Max Response Length      & 2048 \\
Validation Batch Size    & 1024 \\
Random Seed              & 1 \\
\midrule
\multicolumn{2}{l}{\textbf{Optimization}} \\
Global Batch Size (Actor/Critic) & 64 \\
Learning Rate            & $1 \times 10^{-6}$ \\
Weight Decay             & $1 \times 10^{-2}$ \\
Optimizer                & AdamW (bf16) \\
Max Gradient Norm        & 1.0 \\
\midrule
\multicolumn{2}{l}{\textbf{Rollout Configuration}} \\
Number of Rollouts       & 5 (train), 1 (validation) \\
Rollout Temperature      & 1.0 (train), 0.6 (validation) \\
Rollout Top-$p$          & 1.0 (train), 0.95 (validation) \\
Max Batched Tokens       & 10240 \\
\midrule
\multicolumn{2}{l}{\textbf{KL Control}} \\
KL Coefficient ($\lambda_{KL}$) & $1 \times 10^{-2}$ \\
\midrule
\multicolumn{2}{l}{\textbf{Training Setup}} \\
Total Epochs             & 20 (2-entities), 40 (3-/4-entities) \\
\bottomrule
\end{tabular}
}
\label{tab:hyperparameters}
\end{table}

\subsection{Implementation Details}\label{sec:implement_details}

\paragraph{Datasets} Table~\ref{tab:datasets} summarizes the datasets used in our experiments, including their sources, number of entities, and number of triples. 
For FB15k-237, we replace machine identifiers with their corresponding human-readable entity names. 
To address the issue of long and complex relation names in Freebase, we transform each relation into a simplified alias with the assistance of ChatGPT, followed by manual verification. 
For example, the original relation \texttt{/film/actor/film./film/performance/film} is shortened to \texttt{acted\_in}.

\begin{table*}[ht]
\centering
\caption{Statistics of knowledge graph datasets.}
\resizebox{0.82\textwidth}{!}{%
\begin{tabular}{l l r r r}
\toprule
\textbf{Datasets} & \textbf{Source} & \textbf{\#Entities} & \textbf{\#Relations} & \textbf{\#Triples} \\
\midrule
\textbf{Freebase13}   & Freebase & 75,043  & 13  & 375,514 \\
\textbf{FB15k-237}    & Freebase & 14,265  & 237 & 310,116 \\
\textbf{MetaQA}       & Movie-domain & 40,151 & 9   & 134,741 \\
\textbf{DBpedia50}    & Subset of DBpedia (Wikipedia) & 30,449 & 365 & 43,756 \\
\textbf{DBpedia500}   & Larger subset of DBpedia & 490,598 & 573 & 4,268,614 \\
\textbf{YAGO3-10}     & Subset of YAGO3 (Wikipedia, WordNet, GeoNames) & 123,182 & 37 & 1,089,040 \\
\textbf{UMLS}         & Unified Medical Language System & 135 & 46 & 6,529 \\
\bottomrule
\end{tabular}
}

\label{tab:datasets}
\end{table*}

\paragraph{Query Construction} 
For \textit{two-entity} queries, we sampled pairs of seed entities 
such that the shortest path distance $\mathrm{dist}(e_i, e_j) \leq 4$ in knowledge graph. 
The sampling procedure (Algorithm~\ref{alg:sample_pair}) performs repeated random selection of a source entity 
and then selects a target entity reachable within $k$ hops. 
This ensures that queries follow the graph’s local connectivity.

\begin{algorithm}[ht]
\caption{Entity Pair Sampling}
\label{alg:sample_pair}
\begin{algorithmic}[1]
\STATE Randomly sample an entity $u$ from the graph.
\STATE Run BFS up to $k$ hops to obtain reachable set $\mathcal{N}(u,k)$.
\STATE Randomly sample $v \in \mathcal{N}(u,k)$ with $d(u,v)\leq k$.
\STATE Return $(u,v)$ as the entity pair.
\end{algorithmic}
\end{algorithm}

To evaluate scalability, we further constructed queries involving \textit{three} and \textit{four} entities 
on the DBpedia50 dataset. 
For triplets $(u,v,w)$, we required that the third entity $w$ 
is within $k$ hops of at least one of the first two entities ($u$ \textit{or} $v$). 
For quadruplets $(u,v,w,z)$, we extended this requirement so that 
the fourth entity $z$ is within $k$ hops of at least one of $\{u,v,w\}$. 
This OR-connectivity guarantees that multi-entity queries yield semantically well-connected subgraphs.

After sampling seed entities, we converted them into natural language questions 
using a set of manually designed templates. 
For example, given the pair (\textit{Meghan Markle}, \textit{Queen Elizabeth II}), 
we generate the query: 
\textit{``How are Meghan Markle and Queen Elizabeth II associated?''} 
We also used alternative phrasings such as 
\textit{``For what reason are [entity1] and [entity2] related?''} 
to improve linguistic diversity. 

\paragraph{Prompts} For each experiment, we constructed prompts that contextualize the knowledge graph and the query into a unified text input.  
Specifically, the graph is represented as \textbf{node and edge tables}, followed by the natural language query, together with task-specific instructions.  
This design ensures that the model has access to both structural and semantic information.  

An example prompt is shown below.
{\small
\begin{verbatim}
You are given a directed graph as two CSV-like sections in this order:

1) Node table (header included):
node_id, node_attr

2) Edge table (header included):
src, edge_attr, dst

Task
- Use ONLY edges from the Edge table to answer the question by outputting 
  a path.
- When printing each edge, replace IDs with the exact node_attr from the 
  Node table.
- Output MUST be text triples, not numeric IDs.

Output format (STRICT — no extra text):
GRAPH:
("subject"|predicate|"object")
...
END

Rules
- Use only listed edges; do NOT invent edges.
- Map IDs → node_attr; preserve node_attr exactly.
- Output NOTHING outside the SUBGRAPH block.
- If no subgraph exists, output exactly:
GRAPH:
END

Graph:
node_id, node_attr
1, Meghan Markle
2, Prince Harry
3, Prince William
4, Queen Elizabeth II

src, edge_attr, dst
1, spouse, 2
4, grandmother, 2
4, grandmother, 3
2, brother, 3

Question: How are Meghan Markle and Queen Elizabeth II associated?

Your output must be ONLY the SUBGRAPH block.
\end{verbatim}
}

\paragraph{Additional Evaluation Metrics}
Besides the three primary metrics, Connectivity Ratio, Average Reward, and Subgraph F1, 
we report additional reward components to provide a more fine-grained evaluation of relational answers.

\noindent
\textbf{Subgraph F1.}
Subgraph F1 evaluates the overlap between a predicted relational answer and the ground-truth answer subgraph at the triple level.
We define precision and recall over triples, and compute F1 as their harmonic mean:
\begin{equation}
\label{eq:subgraph_f1}
\mathrm{Precision} = \frac{|T_p \cap T_{gt}|}{|T_p|}, \quad
\mathrm{Recall} = \frac{|T_p \cap T_{gt}|}{|T_{gt}|}, \quad
\mathrm{F1} = \frac{2 \cdot \mathrm{Precision} \cdot \mathrm{Recall}}{\mathrm{Precision} + \mathrm{Recall}},
\end{equation}
where $T_p$ and $T_{gt}$ denote the sets of RDF triples in the predicted and ground-truth subgraphs, respectively.

\noindent
In addition to these primary metrics, we report the following reward components:

\squishlist
  \item \textit{Correct Format Percentage} ($F$): the proportion of generated subgraphs that conform to the required output format.
  \item \textit{Average Entity Informativeness} ($I_E$): the \textit{negative} of the average informativeness score of entities (Equation~\ref{eq:entity_informativenss}); lower values indicate more informative entities.
  \item \textit{Average Relation Informativeness} ($I_R$): the \textit{negative} of the average informativeness score of relations (Equation~\ref{eq:relation_informativenss}); lower values indicate more informative relations.
\squishend

\noindent
Note that \textit{Average Entity Informativeness} and \textit{Average Relation Informativeness} are only available when the format reward equals 1 and the generated subgraph is connected.
Therefore, for two-entity queries, we compute these averages only over the connected answers.

\subsection{Details of Main Results}\label{sec:main_results_details}

\paragraph{Parameter Choice.}

\begin{table}[ht]
\centering
\renewcommand{\arraystretch}{1.2}
\caption{Evaluation results of Qwen-2.5-3B-Instruct model on DBpedia500 under different parameter settings. Only reachable cases are reported for $I_R$ and $I_E$.}
\setlength{\tabcolsep}{8pt}
\resizebox{0.4\textwidth}{!}{%
\begin{tabular}{cc c c c c}
\toprule
\textbf{$x$} & \textbf{$y$} & \textbf{$F$} & \textbf{$C$} & \textbf{$I_R$} & \textbf{$I_E$} \\
\midrule
5 & 4 & 100.0\% & 59.0\% & -5.12 & -5.84 \\
6 & 5 & 100.0\% & 60.6\% & -5.62 & -6.43 \\
7 & 6 & 100.0\% & 64.8\%& -5.83 & -7.24 \\
8 & 7 & 100.0\% & 64.2\% & -6.39 & -7.95 \\
\bottomrule
\end{tabular}
}
\label{tab:dbpedia500_results}
\end{table}

For the normalization terms $y$ and $x$ in Equation~\ref{eq:reward}, 
we calibrated their values using the Qwen-2.5-3B-Instruct model on the largest dataset, DBpedia500. 
Since the shortest distance between entity pairs is bounded by $\mathrm{dist}(e_i,e_j) \leq 4$, 
the optimal case for two-entity queries typically involves about four relations and five entities. 
Accordingly, we initialized the parameter search from $(x,y)=(5,4)$.  

Table~\ref{tab:dbpedia500_results} presents the evaluation across different $(x,y)$ values. 
Transitioning from $(5,4)$ to $(6,5)$ leads to a marginal improvement in Connectivity Ratio of 1.6\%, 
but this gain is offset by a reduction in informativeness, with $I_R$ decreasing by 0.50 and $I_E$ by 0.59, 
resulting in a total loss of 1.09. 
By comparison, $(7,6)$ achieves the highest Connectivity Ratio of 64.8\%, representing a 4.2\% increase over $(6,5)$, 
while the informativeness loss remains essentially identical: $I_R$ decreases by 0.21 and $I_E$ by 0.81, 
again summing to 1.09. 
This outcome highlights that $(7,6)$ delivers a substantially better trade-off, 
offering a larger connectivity gain without incurring additional informativeness penalties. 
When further increasing the parameters to $(8,7)$, the Connectivity Ratio ceases to improve and instead begins to decline.  

Based on these observations, we selected $(y,x)=(7,6)$ as the final configuration. 
These values were fixed and applied uniformly across all datasets and models with two-entity queries 
to ensure consistency and comparability.

\paragraph{Performance Comparison across Models and Datasets.}
Table~\ref{tab:main_additional_results} reports the average relation and entity informativeness of Prompting and UniRel across datasets, where lower values indicate more informative relations or entities.

\begin{table*}[ht]
\centering
\caption{Average relation and entity informativeness of Prompting and UniRel where each tuple represents ($I_R$, $I_E$).} 
\resizebox{0.8\textwidth}{!}{%
\begin{tabular}{llccccccc}
\toprule
\multicolumn{2}{c}{\diagbox{\textbf{Models}}{\textbf{Datasets}}} & \textbf{Freebase13} & \textbf{FB15k-237} & \textbf{MetaQA} & \textbf{DBpedia50} & \textbf{DBpedia500} & \textbf{YAGO3-10} & \textbf{UMLS} \\
\midrule
\multirow{2}{*}{\textbf{Qwen3B}}          
& Prompting    & (0.59, 0.67) & (1.93, 1.93) & (1.19, 1.12) & (0.91, 0.53) & (1.31, 1.225) & (0.60, 0.26) & (1.08, 1.61) \\
& UniRel  & (0.34, 0.94) & (2.97, 2.28) & (2.90, 2.45) & (2.01, 1.78) & (3.109, 4.04) & (2.78, 4.69) & (1.25, 4.78) \\ 
\multirow{2}{*}{\textbf{Qwen7B}}          
& Prompting    & (0.32, 0.75) & (1.63, 1.79) & (2.52, 1.63) & (1.07, 0.71) & (2.14, 1.519) & (1.88, 1.43) & (1.10, 2.06) \\
& UniRel  & (0.36, 1.19) & (2.01, 2.84) & (2.90, 2.71) & (2.02, 1.79) & (3.12, 4.20) & (2.81, 4.81) & (1.34, 6.26) \\ 
\multirow{2}{*}{\textbf{Qwen14B}}        
& Prompting    & (0.33, 0.87) & (2.06, 2.52) & (1.93, 0.50) & (1.46, 1.05) & (3.00, 3.00) & (1.82, 1.84) & (1.38, 2.58) \\
& UniRel  & (0.35, 1.20) & (1.69, 4.18) & (2.78, 3.81) & (1.99, 1.98) & (2.99, 5.19) & (2.46, 5.68) & (1.42, 6.71) \\ 
\midrule
\multirow{2}{*}{\textbf{Llama3B}}         
& Prompting    & (-, -) & (-, -) & (-, -) & (-, -) & (-, -) & (-, -) & (-, -) \\
& UniRel  & (0.35, 1.23) & (2.14, 2.87) & (2.14, 2.87) & (1.89, 1.69) & (3.05, 4.22) & (2.89, 4.57) & (0.62, 1.58) \\ 
\multirow{2}{*}{\textbf{Llama8B}}         
& Prompting    & (0.07, 3.07) & (3.87, 6.82) & (-, -) & (-, -) & (-, -) & (4.17, 11.62) & (-, -) \\
& UniRel  & (0.36, 1.29) & (1.97, 4.23) & (1.97, 4.23) & (2.02, 1.96) & (3.03, 5.78) & (2.78, 6.08) & (1.43, 6.83) \\ 
\midrule
\rowcolor{blue!10}
\multicolumn{2}{l}{\textbf{Optimal Search}} 
& (0.29, 0.77) & (1.12, 1.86) & (2.66, 1.78) & (2.08, 1.45) & (2.58, 1.92) & (2.526, 1.67) & (0.91, 1.93) \\
\bottomrule
\end{tabular}
}
\label{tab:main_additional_results}
\end{table*}

For Prompting, the connected answers are typically simple subgraphs with shorter lengths, which explains the relatively low initial scores. After applying UniRel, additional relations and entities may be introduced to ensure connectivity, which can increase the informativeness scores compared to the optimal baseline.

Compared with the optimal search, certain datasets (e.g., Llama-8B on DBpedia50) exhibit lower $I_R$ but higher $I_E$, revealing a gap between relation- and entity-level informativeness. By contrast, datasets with larger reward gaps from the optimal (e.g., FB15k-237) often show a small difference in $I_R$ but a large difference in $I_E$, suggesting that extra high-degree or hub-penalty entities were included to achieve connectivity.

\begin{table*}[ht]
\centering
\caption{Correct format percentage of Prompting and UniRel.} 
\resizebox{0.8\textwidth}{!}{%
\begin{tabular}{llccccccc}
\toprule
\multicolumn{2}{c}{\diagbox{\textbf{Models}}{\textbf{Datasets}}} & \textbf{Freebase13} & \textbf{FB15k-237} & \textbf{MetaQA} & \textbf{DBpedia50} & \textbf{DBpedia500} & \textbf{YAGO3-10} & \textbf{UMLS} \\
\midrule
\multirow{2}{*}{\textbf{Qwen3B}}          
& Prompting   & 83.6\% & 78.8\% & 84.6\% & 84.6\% & 78.2\% & 78.0\% & 93.0\% \\
& UniRel & 100.0\% & 99.8\% & 95.4\% & 100.0\% & 99.8\% & 100.0\% & 100.0\% \\ 
\multirow{2}{*}{\textbf{Qwen7B}}          
& Prompting   & 96.6\% & 93.8\% & 89.4\% & 95.0\% & 90.0\% & 86.8\% & 88.8\% \\
& UniRel & 100.0\% & 100.0\% & 95.8\% & 100.0\% & 100.0\% & 100.0\% & 100.0\% \\ 
\multirow{2}{*}{\textbf{Qwen14B}}        
& Prompting   & 93.4\% & 91.2\% & 94.0\% & 88.2\% & 85.4\% & 94.2\% & 93.8\% \\
& UniRel & 100.0\% & 100.0\% & 95.6\% & 100.0\% & 100.0\% & 100.0\% & 100.0\% \\ 
\midrule
\multirow{2}{*}{\textbf{Llama3B}}         
& Prompting   & 1.6\% & 1.6\% & 4.2\% & 0.4\% & 3.8\% & 3.4\% & 3.8\% \\
& UniRel & 100.0\% & 99.8\% & 99.8\% & 100.0\% & 100.0\% & 100.0\% & 100.0\% \\ 
\multirow{2}{*}{\textbf{Llama8B}}         
& Prompting   & 4.4\% & 12.6\% & 10.2\% & 2.0\% & 16.6\% & 24.6\% & 4.6\% \\
& UniRel & 100.0\% & 100.0\% & 100.0\% & 100.0\% & 100.0\% & 100.0\% & 100.0\% \\ 
\bottomrule
\end{tabular}
}
\label{tab:format_results}
\end{table*}

Table~\ref{tab:format_results} further reveals that Qwen models exhibit a stronger ability to follow prompt instructions under Prompting, maintaining relatively high correct format percentages (often above 80\%).
In contrast, Llama models struggle to produce valid outputs in the absence of structural guidance, with correct format rates frequently below 5\%.
With UniRel, however, both families achieve nearly perfect compliance across all datasets, indicating that reinforcement-guided adaptation is highly effective in enforcing structural answers.

\begin{table*}[ht]
\centering
\caption{Recall and Precision of Prompting and UniRel where each tuple represents (recall, precision).} 
\resizebox{0.8\textwidth}{!}{%
\begin{tabular}{llccccccc}
\toprule
\multicolumn{2}{c}{\diagbox{\textbf{Models}}{\textbf{Datasets}}} & \textbf{Freebase13} & \textbf{FB15k-237} & \textbf{MetaQA} & \textbf{DBpedia50} & \textbf{DBpedia500} & \textbf{YAGO3-10} & \textbf{UMLS} \\
\midrule
\multirow{2}{*}{\textbf{Qwen-3B}}
& Prompting
& (29.9, 29.1) & (0.5, 0.5) & (5.0, 5.0) & (14.9, 17.6)
& (2.9, 4.9) & (3.9, 4.4) & (13.9, 10.8) \\
& UniRel
& (59.3, 58.2) & (3.6, 2.2) & (25.8, 24.4) & (39.7, 38.8)
& (15.4, 13.7) & (16.6, 15.2) & (17.3, 17.0) \\

\multirow{2}{*}{\textbf{Qwen-7B}}
& Prompting
& (47.6, 40.1) & (0.9, 1.0) & (8.6, 13.4) & (24.0, 29.0)
& (6.5, 10.7) & (8.0, 10.4) & (15.7, 12.1) \\
& UniRel
& (60.8, 59.8) & (3.7, 3.2) & (27.0, 26.4) & (41.0, 40.0)
& (17.1, 15.3) & (18.5, 17.3) & (17.7, 17.6) \\

\multirow{2}{*}{\textbf{Qwen-14B}}
& Prompting
& (54.3, 43.6) & (2.0, 1.9) & (14.1, 20.4) & (30.4, 36.8)
& (9.3, 13.0) & (11.3, 12.7) & (15.7, 12.1) \\
& UniRel
& (63.7, 61.1) & (4.1, 3.5) & (28.4, 27.9) & (44.2, 43.5)
& (17.2, 16.7) & (18.3, 17.7) & (16.7, 16.7) \\

\midrule
\multirow{2}{*}{\textbf{Llama-3B}}
& Prompting
& (0.0, 0.0) & (0.0, 0.0) & (0.0, 0.0) & (0.0, 0.0)
& (0.0, 0.0) & (0.0, 0.0) & (0.0, 0.0) \\
& UniRel
& (62.6, 59.8) & (3.4, 2.6) & (28.0, 26.1) & (27.1, 26.5)
& (13.4, 12.5) & (14.3, 14.3) & (15.4, 15.6) \\

\multirow{2}{*}{\textbf{Llama-8B}}
& Prompting
& (0.0, 0.0) & (0.0, 0.0) & (0.0, 0.0) & (0.0, 0.0)
& (0.0, 0.0) & (0.0, 0.0) & (0.0, 0.0) \\
& UniRel
& (65.7, 64.9) & (4.6, 3.8) & (27.5, 26.8) & (43.2, 42.2)
& (18.6, 17.6) & (19.2, 18.5) & (17.9, 17.8) \\

\bottomrule
\end{tabular}
}
\label{tab:main_additional_results_recall}
\end{table*}

Table~\ref{tab:main_additional_results_recall} reports the triple-level recall and precision of Prompting and UniRel across multiple knowledge graph benchmarks.
Across the evaluated benchmarks, UniRel generally achieves higher recall and precision than Prompting.
Notably, the gains are not limited to recall alone. UniRel also achieves higher precision, suggesting that the recovered triples are not only more complete but also more accurate, rather than being obtained through indiscriminate expansion.
Furthermore, the improvements persist as model size increases, indicating that UniRel complements model scaling by providing better structural guidance instead of relying solely on increased model capacity.

\paragraph{Case Study.}  
We present three additional case studies to further illustrate the differences between the smallest subgraph and the UniRel.  

\begin{figure*}[ht]
    \centering
    \includegraphics[width=0.95\textwidth]{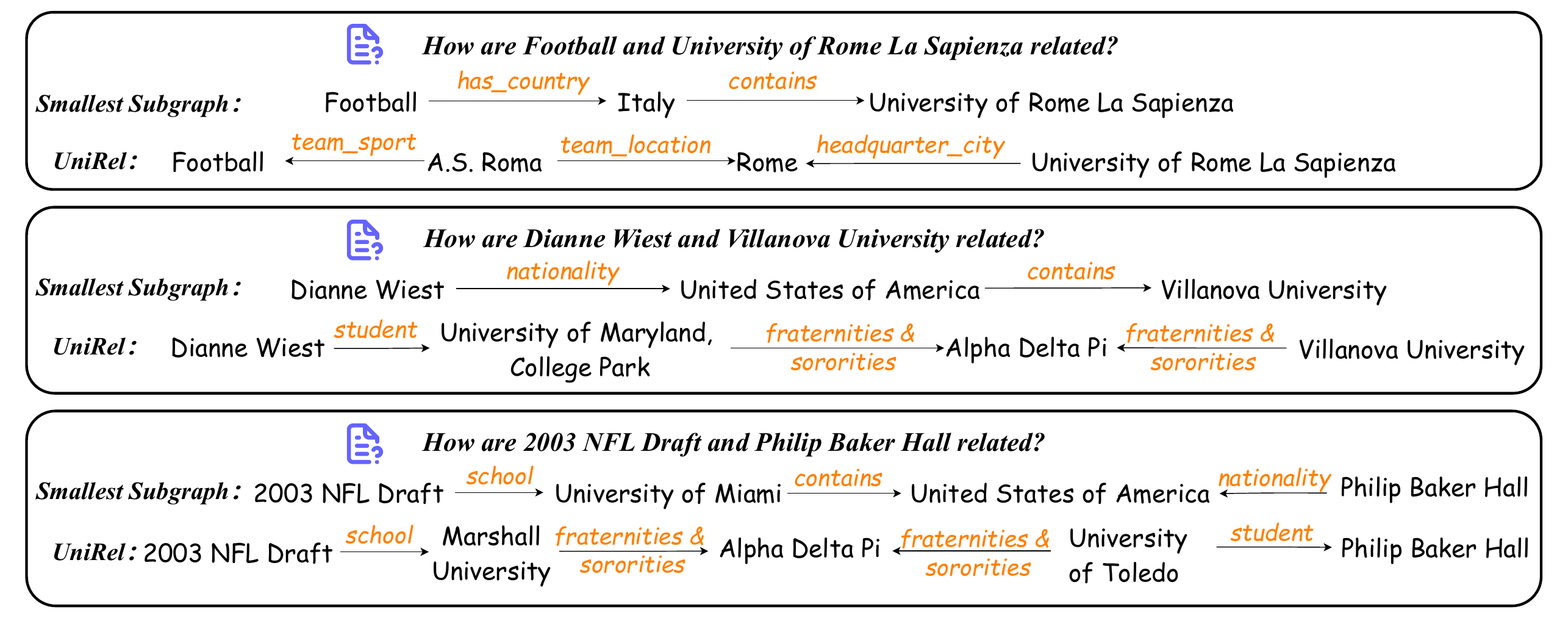} 
    \label{fig:case_studies_2}
\vspace{-1em}
\end{figure*}

\subsection{Details of Generalization Results}\label{sec:generalization_results_details}

Table~\ref{tab:generalization_results_complete} reports the complete cross-dataset generalization results of UniRel trained on DBpedia500 and evaluated on all benchmarks.
Beyond the partial results discussed in the main paper, we provide additional analysis to better understand the observed performance variations across datasets.

We observe that generalization performance is relatively strong on DBpedia50, which may be attributed to its high similarity to the training dataset, DBpedia500, in terms of data source, schema, and graph structure.
Both datasets are derived from DBpedia and share similar entity types, relation distributions, and structural patterns, which likely facilitates knowledge transfer under the UniRel framework.

In contrast, generalization performance on UMLS is comparatively weaker.
A possible explanation is the substantial semantic gap between UMLS and DBpedia-based datasets.
UMLS focuses on biomedical concepts and relations that differ significantly from the general-domain entities and relations seen during training.
As a result, semantic similarity learned during pretraining and fine-tuning may be less effective in supporting cross-dataset transfer, limiting the benefits of semantic generalization.

Taken together, these observations suggest that cross-dataset generalization under UniRel is influenced by both semantic similarity and structural compatibility between the training and target knowledge graphs.
Datasets that are closer to the training domain in terms of semantics and structure tend to exhibit stronger transfer performance, while datasets with larger semantic shifts pose greater challenges.

\begin{table*}[ht]
\centering
\caption{Cross-dataset generalization of UniRel trained on DBpedia500, reported as (connectivity ratio, average reward, subgraph F1).}
\resizebox{0.9\textwidth}{!}{%
\begin{tabular}{llccccc}
\toprule
\multicolumn{2}{c}{\diagbox{\textbf{Datasets}}{\textbf{Models}}} & \textbf{Qwen3B} & \textbf{Qwen7B} & \textbf{Qwen14B} & \textbf{Llama3B} & \textbf{Llama8B} \\
\midrule
\multirow{2}{*}{\textbf{Freebase13}} 
 & \cellcolor{gray!12}Original & (80.0\%, 0.52, 44.7) & (83.4\%, 0.59, 46.1) & (94.0\%, 0.77, 52.7) & (81.8\%, 0.55, 46.9) & (91.0\%, 0.72, 54.1) \\
 & \cellcolor{yellow!12}Modified & (69.6\%, 0.33, 41.3) & (74.8\%, 0.43, 44.2) & (84.0\%, 0.60, 50.4) & (81.6\%, 0.55, 46.9) & (87.0\%, 0.65, 51.2) \\
\midrule
\multirow{2}{*}{\textbf{FB15k-237}}  
 & \cellcolor{gray!12}Original & (47.0\%, -0.23, 1.8) & (61.6\%, 0.01, 1.7) & (76.6\%, 0.16, 3.1) & (45.6\%, -0.28, 1.6) & (69.8\%, 0.08, 2.8) \\
 & \cellcolor{yellow!12}Modified & (20.8\%, -0.71, 0.4) & (41.4\%, -0.32, 1.2) & (52.4\%, -0.12, 1.8) & (45.8\%, -0.25, 1.7) & (59.4\%, -0.03, 1.9) \\
\midrule
\multirow{2}{*}{\textbf{MetaQA}}     
 & \cellcolor{gray!12}Original & (60.6\%, -0.12, 20.0) & (69.8\%, 0.04, 21.7) & (78.0\%, 0.11, 23.1) & (64.2\%, -0.06, 21.7) & (80.0\%, 0.18, 23.9) \\
 & \cellcolor{yellow!12}Modified & (29.2\%, -0.56, 9.1)  & (49.0\%, -0.21, 16.5)  & (60.8\%, -0.02, 18.4)  & (67.2\%, 0.06, 22.5) & (72.6\%, 0.15, 23.2) \\
\midrule
\multirow{2}{*}{\textbf{DBpedia50}}  
 & \cellcolor{gray!12}Original & (91.4\%, 0.55, 40.1) & (96.6\%, 0.64, 42.1) & (97.6\%, 0.66, 42.3) & (56.8\%, -0.02, 34.8) & (98.2\%, 0.66, 42.5) \\
 & \cellcolor{yellow!12}Modified & (54.6\%, -0.06, 34.0) & (73.0\%, 0.27, 37.9) & (84.0\%, 0.45, 40.7) & (84.6\%, 0.45, 38.8) & (91.8\%, 0.57, 41.7) \\
\midrule
\multirow{2}{*}{\textbf{YAGO3-10}}   
 & \cellcolor{gray!12}Original & (56.6\%, -0.18, 11.5) & (67.4\%, 0.01, 13.2) & (80.6\%, 0.15, 15.0) & (44.8\%, -0.36, 10.1) & (83.0\%, 0.18, 15.8) \\
 & \cellcolor{yellow!12}Modified & (23.6\%, -0.66, 7.23) & (41.4\%, -0.34, 10.0) & (48.2\%, -0.24, 11.4) & (44.6\%, -0.31, 10.5) & (62.2\%, -0.03, 13.8) \\
\midrule
\multirow{2}{*}{\textbf{UMLS}}       
 & \cellcolor{gray!12}  Original & (49.0\%, -0.19, 13.3) & (55.8\%, -0.07, 13.6) & (74.0\%, 0.10, 12.7)          & (56.4\%, -0.12, 13.4) & (66.6\%, 0.05, 13.1) \\
 \multirow{0}{*}{}
 &\cellcolor{yellow!12} Modified & (41.0\%, -0.30, 13.1) & (43.4\%, -0.24, 13.4) & (50.0\%, -0.14, 12.1) & (49.4\%, -0.17, 13.6) & (55.0\%, -0.07, 12.4) \\
\bottomrule
\end{tabular}
}
\label{tab:generalization_results_complete}
\end{table*}

\subsection{Details of UniRel for Entity-Centric KGQA}\label{sec:entity_centric_results_details}

\paragraph{Datasets.}

We evaluate UniRel on two widely used multi-hop KGQA benchmarks, WebQSP and CWQ as shown in Table~\ref{tab:dataset_stats}.
Both datasets are built on Freebase and require reasoning over multiple entities and relations, making them well suited for evaluating relation-centric and subgraph-based reasoning methods.

WebQSP focuses on relatively constrained multi-hop questions, where answers can often be derived from compact subgraphs connecting a small number of entities.
In contrast, CWQ contains more complex questions with longer reasoning chains and more diverse relational structures, posing greater challenges for subgraph extraction and relational reasoning.

\begin{table}[ht]
\centering
\small
\caption{Dataset statistics.}
\begin{tabular}{lccc}
\toprule
\textbf{Dataset} & \textbf{\#Train} & \textbf{\#Test} & \textbf{Max Hop} \\
\midrule
WebQSP & 2{,}826 & 1{,}628 & 2 \\
CWQ    & 27{,}639 & 3{,}531 & 4 \\
\bottomrule
\end{tabular}
\label{tab:dataset_stats}
\end{table}

\paragraph{Metrics.}
\textbf{F1}: measures the overlap between the predicted answer set and the ground-truth answer set by combining precision and recall.
It reflects how well the model balances answer correctness and completeness, and is particularly suitable for questions with multiple correct answers.

\textbf{Hit@1}: measures whether the top-ranked predicted answer exactly matches the ground-truth answer.
It captures the model’s ability to place the correct answer at the highest rank and is commonly used for single-answer or ranking-based evaluation.

%%%%%%%%%%%%%%%%%%%%%%%%%%%%%%%%%%%%%%%%
% \newpage
% \appendix


% You can have as much text here as you want. The main body must be at most $8$
% pages long. For the final version, one more page can be added. If you want, you
% can use an appendix like this one.

% The $\mathtt{\backslash onecolumn}$ command above can be kept in place if you
% prefer a one-column appendix, or can be removed if you prefer a two-column
% appendix.  Apart from this possible change, the style (font size, spacing,
% margins, page numbering, etc.) should be kept the same as the main body.
%%%%%%%%%%%%%%%%%%%%%%%%%%%%%%%%%%%%%%%%%%%%%%%%%%%%%%%%%%%%%%%%%%%%%%%%%%%%%%%
%%%%%%%%%%%%%%%%%%%%%%%%%%%%%%%%%%%%%%%%%%%%%%%%%%%%%%%%%%%%%%%%%%%%%%%%%%%%%%%

\end{document}